\documentclass{article}

\usepackage{parskip}
\usepackage{enumitem}
\usepackage[numbers]{natbib}
\usepackage[english]{babel}
\usepackage{graphicx}
\usepackage{caption}
\usepackage{subcaption}
\usepackage{amsmath, amssymb, mathtools, amsfonts}
\usepackage{multirow}

\usepackage{float}
\usepackage{algorithm}
\usepackage{algorithmic}

\usepackage{xcolor} 
\usepackage{soul} 

\usepackage[colorinlistoftodos]{todonotes}

\usepackage[verbose=true,letterpaper]{geometry}
\AtBeginDocument{
  \newgeometry{
    textheight=9in,
    textwidth=5.5in,
    top=1in,
    headheight=12pt,
    headsep=25pt,
    footskip=30pt
  }
}

\usepackage[]{hyperref}         
\hypersetup{                    
  colorlinks,
  linkcolor={blue!80!black},
  citecolor={blue!70!black},
  urlcolor={blue!70!black}
}

\newcommand{\real}{\mathbb{R}}
\newcommand{\imp}{\hat{\Phi}}

\newcommand{\trueimp}{\Phi}
\newcommand{\basevalue}{v}
\newcommand{\basematrix}{\mathbf{v}}
\newcommand{\features}{[p]}
\newcommand{\sampleset}{\mathcal{D}}
\newcommand{\truetopk}{\mathcal{T}_k}
\newcommand{\topk}{\widehat{\mathcal{T}}_k}
\newcommand{\paranking}{D}

\newtheorem{theorem}{Theorem}

\newtheorem{definition}{Definition}

\newtheorem{lemma}{Lemma}

\title{Confident Feature Ranking}

\author{
  Bitya Neuhof\\
  Department of Statistics and Data Science\\
  The Hebrew University of Jerusalem\\
  \texttt{bitya.neuhof@mail.huji.ac.il}\\
  \and
  Yuval Benjamini\\
  Department of Statistics and Data Science\\
  The Hebrew University of Jerusalem\\
  \texttt{yuval.benjamini@mail.huji.ac.il}
}

\begin{document}

\maketitle

\begin{abstract}
\normalsize
    Machine learning models are widely applied in various fields. Stakeholders often use post-hoc feature importance methods to better understand the input features' contribution to the models' predictions. The interpretation of the importance values provided by these methods is frequently based on the relative order of the features (their ranking) rather than the importance values themselves. Since the order may be unstable, we present a framework for quantifying the uncertainty in global importance values. We propose a novel method for the post-hoc interpretation of feature importance values that is based on the framework and pairwise comparisons of the feature importance values. This method produces simultaneous confidence intervals for the features' ranks, which include the ``true'' (infinite sample) ranks with high probability, and enables the selection of the set of the top-k important features.
\end{abstract}

\section{Introduction}\label{sec:introduction}

Complex nonlinear prediction models are widely used to augment or even replace human judgement in fields such as healthcare \citep{bhardwaj2017study}, finance \citep{rundo2019machine}, and science \citep{deiana2022applications, li2022machine}. Regulators, users, and developers of such models are interested in understanding the relative contribution of the different inputs, i.e., features, to the model's predictions \citep{preece2018stakeholders, goodman2017european}.
Feature importance (FI) methods such as permutation feature importance (PFI) \citep{breiman2001random} and SHapley Additive exPlanations (SHAP) \citep{lundberg2017SHAP, lundberg2019Tree} measure the contribution of features by estimating the effect of removing, perturbing, or permuting the feature on the predicted value or prediction loss. The specifics of this manipulation vary depending on the method and implementation \citep{merrick2020explanation, covert2020imputation}.
These FI methods are employed to explain the predictions of models after they  have been trained, and therefore they are called \emph{post-hoc FI methods}. This paper focuses on global FI methods that explain the average model behavior rather than local FI methods that explain individual predictions.

Recently, studies have demonstrated that post-hoc FI methods can be unstable \citep{molnar2020general, marx2023but} due to uncertainty stemming from the size and sampling of the data used to calculate the FI values (explanation set); randomness in the perturbations, permutations \citep{lakkaraju2020robust, agarwal2022rethinking}, or approximations \citep{merrick2020explanation};  hyperparameter selection \citep{slack2021reliable, ahn2023local}; and more. We focus on uncertainty in sampling the explanation set, which affects the stability of the FI values. Most methods for quantifying this type of uncertainty produce per-feature spread estimates (or confidence intervals) in the FI method's output units \citep{ishwaran2019standard, covert2020SAGE, merrick2020explanation, slack2021reliable, ahn2023local, molnar2021relating}.

\begin{figure}[ht]
     \centering
     \includegraphics[width=0.49\columnwidth]{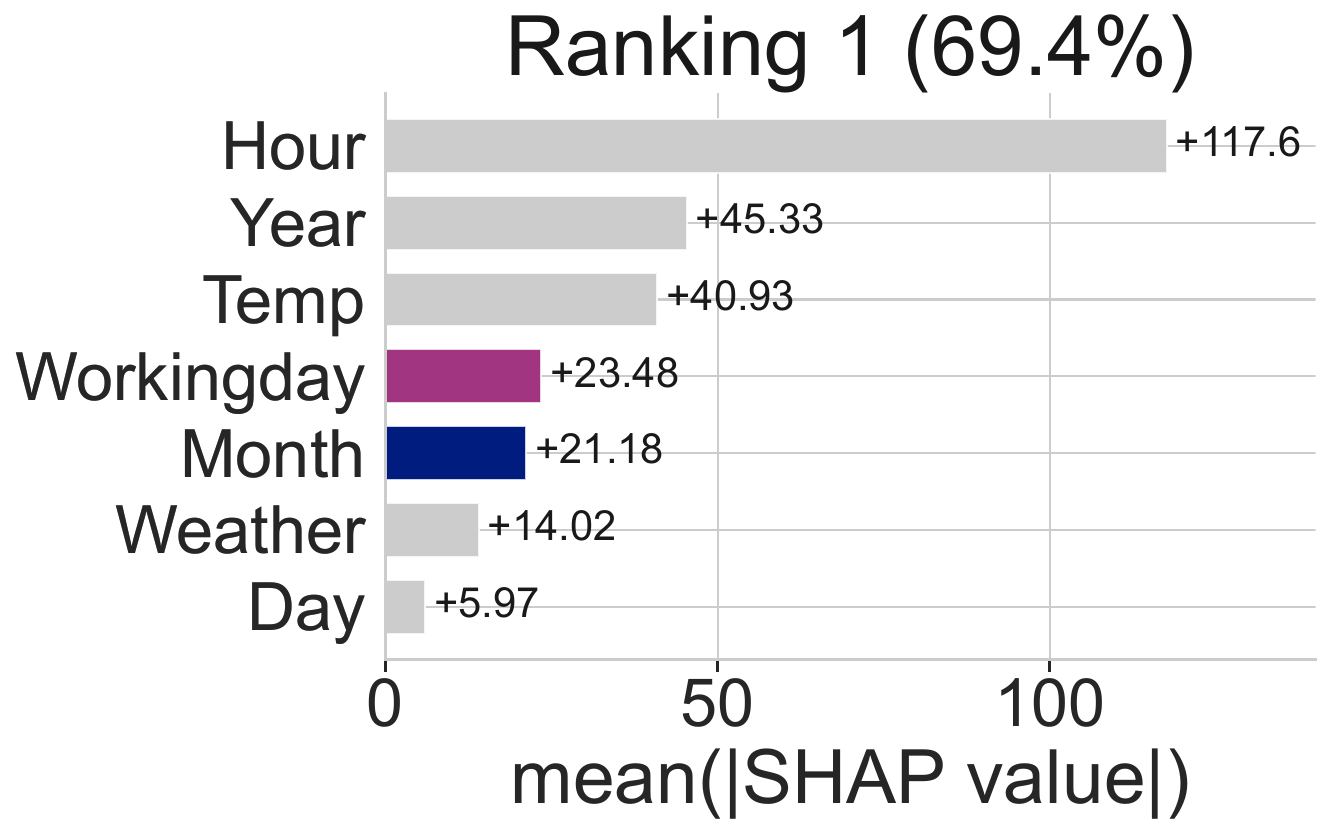}
     \includegraphics[width=0.49\columnwidth]{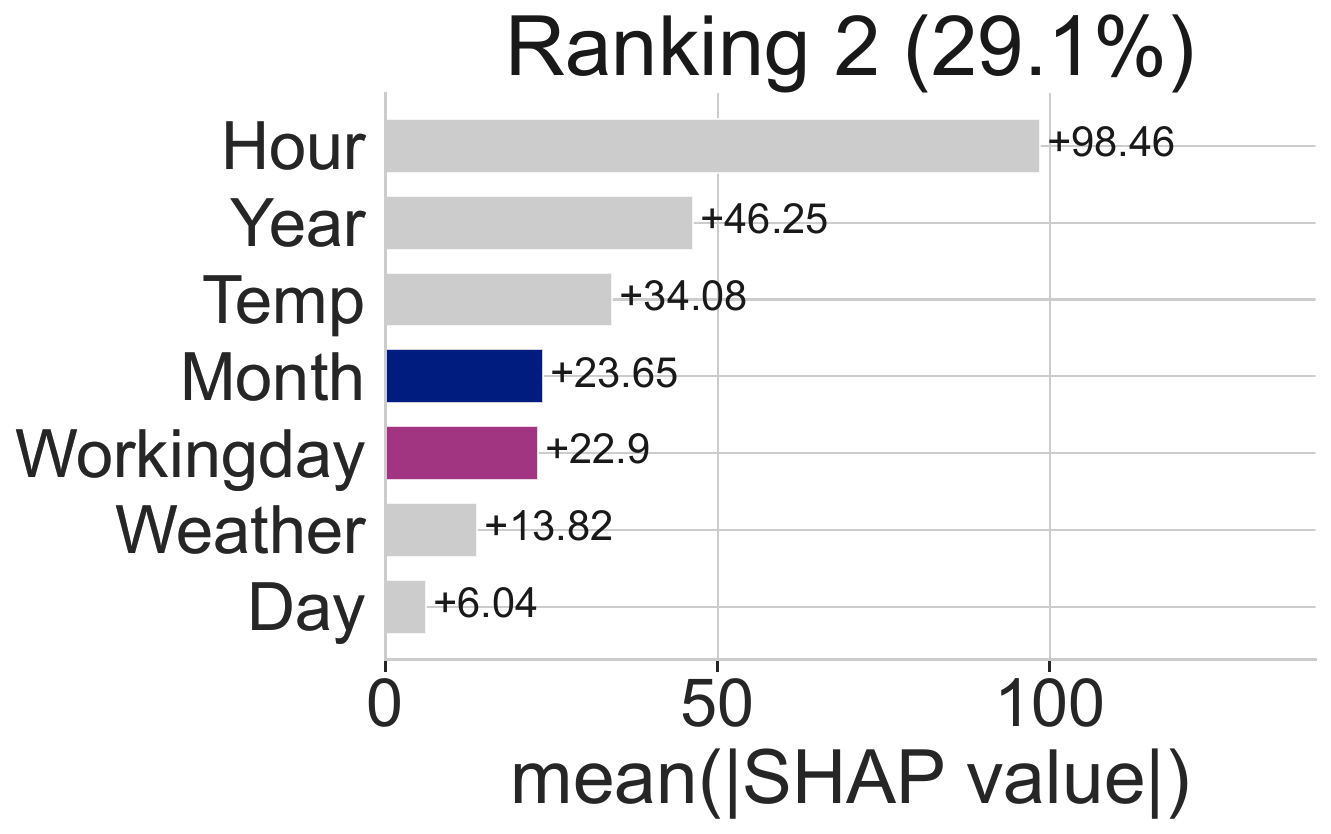}
     \caption[]%
     {Bar plots of SHAP values for two samples of $n=50$ observations from the bike sharing dataset using an XGBoost model. The ranking of the features is unstable for this sample size: ranking of the \emph{Workingday} and \emph{Month} features varies, depending on the sample. The chances of observing each of the rankings (69.4\% and 29.1\%) are estimated based on 1,000 independent samples of size $n=50$.}
     \label{fig:intro}
\end{figure}

Existing uncertainty measures are insufficient, because stakeholders often rely on the \emph{rank of the FI value}, rather than the value itself, in their decisions. Feature rankings are unit-independent and are therefore easy to interpret and compare across FI methods \citep{jaxa2021sources, heldt2021early}. Instability in the global FI values can lead to instability in their ranking \citep{rising2021uncertainty} (an example is provided in Figure \ref{fig:intro}). A simple ranking of the features based on the FI values cannot reflect this uncertainty. Moreover, due to the ranking's discrete nature, existing methods for quantifying uncertainty in FI values cannot easily be modified to work for ranking uncertainty. For example, we show that confidence intervals (CIs) produced by a naive bootstrapping method based on the estimation of the ranking distribution do not cover the true ranks. The previously mentioned challenges point to the need for a framework for defining, estimating, and reporting ranking uncertainty. To properly model ranking uncertainty, we first model the uncertainty of the global FI values and then infer the effect of this uncertainty on the rankings.

In this paper, we present a \emph{base-to-global} framework to quantify the uncertainty of global FI values. We define a two-level hierarchy of importance values, namely the \emph{base} and \emph{global} FI values, where the global FI values are the average of independent base FI values.
Based on this framework, we propose a novel method for confidently ranking features. We define the \emph{true rank} as a feature's rank, obtained based on an infinite sample, for both a trained prediction model and an FI method. Our ranking method reports simultaneous CIs, ensuring, with high probability, that each feature's true rank is covered by the appropriate interval. We construct the intervals by examining all pairs of features, testing hypotheses regarding differences in means, and counting the number of rejections for each feature. The examination process tackles the multiple tests problem, which might result in the false discovery of a feature as relevant. The validity of our proposed method is demonstrated in a comprehensive evaluation on both synthetic and real-world datasets. Our findings confirm our method's effectiveness and highlight its potential in quantifying and enhancing ranking stability. Our base-to-global framework can be viewed as a generalization of the formulate, approximate, explain (FAE) \citep{merrick2020explanation} framework for generating and interpreting Shapley-value-based FI methods. We extend the FAE concept in two respects: first, we generalize it to other post-hoc FI methods by defining the base values in a general way; and second, we address the uncertainty in the ranking of the global FI values.

Our main contributions in this paper are as follows: (1) We propose a novel ranking method for FI values. (2) We quantify the uncertainty of the ranking by providing simultaneous CIs for the features' ranks.\footnote{The paper's code is publicly available at: \url{https://github.com/BityaNeuhof/confident_feature_ranking}.} (3) We suggest an improved means of interpreting global FI values.
We generalize confident ranking methods to accommodate correlations and potential departures from normality, which are common in FI values. To the best of our knowledge, our ranking method is the first to formally incorporate uncertainty control in the \emph{ranking} of FI values.
\section{Quantifying Uncertainty in Global FI Values}\label{sec:framework}

\subsection{Terminology}

Consider the supervised learning task of predicting a real-value outcome $Y \in \mathcal{Y}$ from a vector of $p$ features $X = (X_1,\ldots,X_p) \in \mathcal{X}$.
A prediction model $f: \mathcal{X} \rightarrow \mathcal{Y}$ is trained on a training set 
$\sampleset_{train} =  \{(x_i, y_i)\}_{i=1}^M$ and fits the data well according to standard metrics (e.g., MSE or accuracy on external test sets). 
Stakeholders are then interested in the extent to which a feature contributes to the model's performance or predictions -- the FI value.

\subsection{Base-to-Global Framework}

Post-hoc global FI methods describe the average behavior of the model. These methods produce an importance value for each feature, $\imp_1, \imp_2, \ldots, \imp_p \in \real$, based on a trained model $f$ and a sample $\sampleset_{explain} = \{(x_i, y_i)\}_{i=1}^N$, preferably independent of $\sampleset_{train}$. In most methods, the assumption is that a higher value of $\imp_j$ indicates greater importance. Generally, the features are ranked according to their FI values, and only the top-$k$ features are considered.

In many cases, the FI values are calculated by averaging many independent runs. For example, in SHAP \citep{lundberg2017SHAP}, the global FI value is an average of the absolute values assigned to each observation (the local SHAP values). Variability in the explanation set $\sampleset_{explain}$ introduces uncertainty into the global FI values.\footnote{This paper only considers the exact computation of SHAP values without approximation.} In PFI \citep{breiman2001random}, the global FI value is the average obtained over multiple permutations. In this case, both variability in the explanation set and the randomized permutations introduce uncertainty into the global FI values.

In considering how these examples could be addressed in a single framework, we make the following observation: there is a two-level FI hierarchy in which the observed \emph{global} FI value is an average of independent \emph{base} FI values; in the first example (SHAP), the base FI values correspond to the local SHAP values, and in the second example (PFI), the base FI values correspond to the PFI values calculated for a single permutation on the full explanation set.

We set the following notations: matrix $\basematrix_{n \times p}$ is defined as the matrix of \emph{base} FI values,
with rows  $\basevalue_1, \ldots, \basevalue_n \in \real^p$ representing the FI value for each feature.\footnote{If the base FI values are the local values, $n = N$ is the size of $\sampleset_{explain}$.}
$\basematrix_j$ are the columns of the matrix referring to the base FI values for the $j$'th feature. The observed global FI value for the $j$'th feature is $\imp_j =\frac{1}{n} \sum \basevalue_{ij}$.

\paragraph{SHAP Example} For a single observation $(x, y)$, the local SHAP value of a feature $j$ is:
\begin{align*}
    &\phi_j = \sum_{S \subseteq \features \setminus \{j\}} \frac{|S|!(p - |S|-1)!}{p!} \times \\
    &\Big(\mathbb{E}[f(X)|X_{S \cup \{j\}}=x_{S \cup \{j\}}]
    - \mathbb{E}[f(X)|X_S=x_S]\Big)
\end{align*}

where $\features$ is the set of all features, and $S$ is a subset of features.
The base FI value is:
\begin{equation}\label{eq:local_shap}
    \basevalue_j^{SHAP} = |\phi_j|,
\end{equation}
and the global FI value is: $\imp_j^{SHAP} =  \frac{1}{n}\sum_{i=1}^n \basevalue_j^{SHAP}$.

\paragraph{PFI Example} Let $L$ be a loss function; the global PFI value of a feature $j$ is:
\begin{align*}
    \imp_j^{PFI}=\frac{1}{B}\sum_{b=1}^B L(f(X_{[j]}^b), Y) - L(f(X), Y),
\end{align*}
where $X_{[j]}$ is a replication of $X$ with a permuted version of the $j$'th feature, and $B$ is the number of permutations. The base FI value can be defined either as a single permutation of the $j$'th feature: $\basevalue_j^{PFI}=L(f(X_{[j]}), Y) - L(f(X), Y)$ (here the number of base FI values is the number of permutations ($n=B$)) or as the average of permutations for an observation:
\begin{equation}\label{eq:local_pfi}
    \basevalue_j^{PFI} = \frac{1}{B}\sum_{b=1}^B
    L(f(x_{[j]}^b), y) - L(f(x), y),
\end{equation}
where $x_{[j]}$ is a replication of observation $x$ with a permuted version of the $j$'th feature. Here, the number of base FI values is the number of observations ($n=N$). A detailed analysis of the sources of uncertainty in PFI is provided in Appendix \ref{app:pfi_var}.

\subsection{Uncertainty in Feature Ranking}

Global FI values are often interpreted as a ranking used to highlight or select the most relevant features. Since different FI methods produce FI values of varying scales, the ranking of the features is often used to compare the methods' output. The observed ranks $\hat{r} = (\hat{r}_1, \ldots, \hat{r}_p)$, $\hat{r}_j \in \{1,...,p\}$ are typically derived directly from the observed global FI values, with the rank $p$ assigned to the highest global FI value, and rank $1$ assigned to the lowest.

\begin{figure}[ht]
     \centering     \includegraphics[width=0.8\columnwidth]{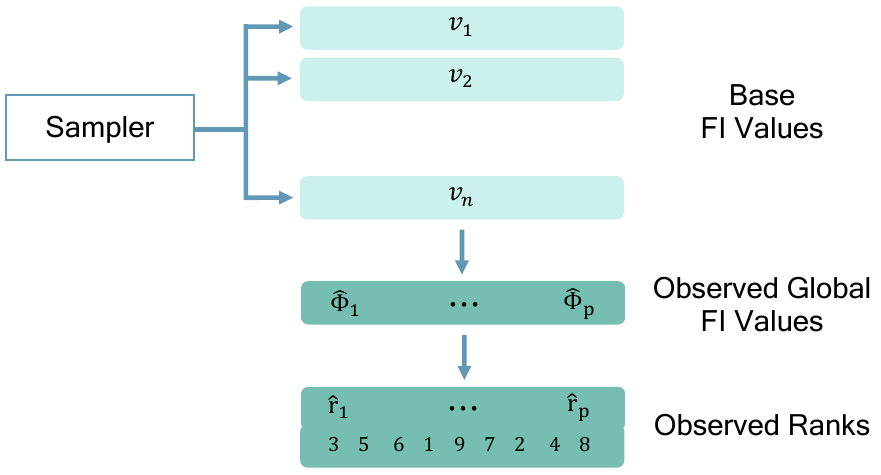}
     \caption[]%
     {Base FI values are sampled as vectors, introducing uncertainty. The vectors are averaged to form the observed global FI values. Finally, the global FI values are ranked to produce the observed ranks.}
     \label{fig:framework}
\end{figure}

The sampling of the base FI values introduces uncertainty into the global FI values. The global FI values are then ranked, propagating the uncertainty into the observed ranks. This process is summarized in Figure \ref{fig:framework} which presents the framework's pipeline for quantifying the uncertainty in the observed ranks.

Note that the definitions of the base and global FI levels specify the source of the uncertainty to be reflected in the CIs for the ranks; two options for base FI value definition for PFI are presented above.
\section{Statistical Model and Inference Goal}\label{sec:prob}

In this section, we describe our statistical model for estimating ranking uncertainty. First, we define a feature's rank-set as a rank that considers information about ties. Then we define our inference goal -- to provide simultaneous CIs for the rank-sets. Finally, we discuss the advantages of simultaneous CIs and provide an example in which the top-$k$ features are highlighted.

\subsection{True Global FI Values and Rank-Sets}

Recall that in Section \ref{sec:framework} we introduced the base FI values matrix $\basematrix$, with rows $\basevalue_i \in \real_p$. Here, we model the rows $\basevalue_i$ as independent samples from distribution $F_\basevalue$ with mean vector $E[\basevalue_i] = (\trueimp_1,\ldots,\trueimp_p)$; these are the \emph{true global FI values}. Each observed global FI value $\imp_j$ is an unbiased but noisy version of $\trueimp_j$. We are interested in understanding the effects of this variability on the possible feature rankings.

In contrast to the observed noisy ranks, the \emph{true ranks} $r_1,...,r_p$ are based on the true global FI values $\trueimp_1,...,\trueimp_p$.
Whereas in the observed global FI values exact ties are unlikely, for the true global FI values we can imagine ties between equivalent features, or we may want to allow an indifference level. We follow \cite{al2022simultaneous} in redefining the true ranks to account for ties:
\begin{definition}(Rank-Set)
\label{def:setrank}
    Define the lower rank of $\trueimp_j$ as $
        l_j = 1 + \#\{k: \trueimp_j > \trueimp_k, \, j \neq k\}$
    and the upper rank of $\trueimp_j$ as $
        u_j = p - \#\{k: \trueimp_j < \trueimp_k, \, j \neq k\} $.    Then the rank-set of $\trueimp_j$ is the set of natural numbers $\{l_j, l_{j+1}, \ldots, u_j\}$ denoted as $[l_j, u_j]$.
\end{definition}

If there are no ties, the lower and upper ranks are identical and equal to the standard definition. In the remainder of the paper, the term `true rank' will refer to the rank-set.

\subsection{CIs for True Ranks}

We propose quantifying the rankings' uncertainty using simultaneous CIs for the true ranks. Here, we define marginal and simultaneous CIs, and in Section \ref{sec:conf_rank} we propose a method for constructing valid simultaneous CIs for the true ranks.

\begin{definition}(CI for a True Rank)
    The ranks interval $[L_j, U_j]$ is an $(1-\alpha)$-level CI for a true rank of the $j$'th feature if $\mathbb{P}_{F_\basevalue}\left([l_j, u_j] \subseteq [L_j, U_j]\right) \geq 1 - \alpha$ for any possible $F_\basevalue$.
\end{definition}

$L_j, U_j$ are functions of $\basematrix$, the matrix of observed base FI values. We note that different sets of observed base FI values will produce different CIs.

The set of intervals $[L_1,U_1] , \ldots, [L_p, U_p]$ has marginal coverage if each interval is a valid CI of the corresponding true rank. For ranking and selection of features, marginal coverage is not sufficient, because it does not support selection after ranking \citep{benjamini2005false}. Therefore, our ranking method constructs simultaneous CIs for the true ranks.

\begin{definition}(Simultaneous Coverage) The set of intervals $[L_1,U_1] , \ldots, [L_p, U_p] \subseteq [1, p]$ has simultaneous coverage at level $1 - \alpha$
if 
\begin{align*}
    \mathbb{P}_{F_\basevalue}\left([l_j, u_j] \subseteq [L_j, U_j] , \quad \forall j \in \{1, \ldots, p\}\right) \geq 1 - \alpha.
    \end{align*}
\end{definition}

In $1-\alpha$ simultaneous CIs, the probability that all intervals cover the true ranks is at least $1-\alpha$. Simultaneous CIs remain valid for any form of selection after ranking (for example, selection of the most important features). We note that simultaneous coverage is conservative and can result in relatively large intervals.

\subsection{Top-K Set}\label{sec:top-k}

Here we present an application of simultaneous CIs for the selection of the most important features (top-$k$) with a guarantee of coverage. Since the ranking is based on the observed FI values, the size of the set of possible top-$k$ features might be greater than $k$.

Denote $\truetopk \subseteq \features$ as the set of features whose true FI value is ranked in the top-$k$ $\truetopk = \{j: u_k \geq p-k+1 \}$. A simple selection method is to select features for which the upper bound of the CI is greater than $p - k$. With simultaneous coverage, the probability of an error for this selection is controlled \citep{hsu1996multiple}. Furthermore, the CIs for the features currently ranked among the top-$k$ still have marginal coverage. These two properties are not guaranteed without simultaneous coverage \citep{ein2006thousands}.

\begin{lemma}
Let $\{ [L_1, U_1],\ldots,[L_p,U_p] \}$ be $1-\alpha$ simultaneous CIs for the true ranks. Define the \emph{top-k set} $\topk  = \{j: U_j \geq p-k+1 \}$. This set includes all features with an upper bound in the top-$k$ ranks $(p, p - 1, \ldots, p - k + 1)$. Then $\mathbb{P}(\truetopk \subseteq \topk) \geq 1-\alpha$.
\end{lemma}

To prove this, consider a case in which $\truetopk \subseteq \topk$ does not hold; then it must follow that there is some $j\in \truetopk$ that is not in $\topk$. This means that the estimated upper bound $U_j$ is less than the true upper rank $u_j$, so the CI $[L_j,U_j]$ does not cover $[l_j, u_j]$. Based on the definition of $1-\alpha$ simultaneous CIs, the probability of any such event is at most $\alpha.$
\section{Confident Simultaneous Feature Ranking}\label{sec:conf_rank}

In this section, we introduce our ranking method which is designed to rank FI values while taking into account the uncertainty associated with the post-hoc FI method and the sampling process. Using our base-to-global framework, we are able to quantify the uncertainty by calculating simultaneous CIs for the true ranks.

\subsection{Feature Ranking}

Our method uses pairwise hypothesis tests to estimate lower and upper bounds for the true rank of each feature.
For each feature pair $j,k$, we perform two one-sided hypothesis tests:
\begin{enumerate}
\item[(a)] A test of
$H^1_{jk}:  \trueimp_j < \trueimp_k \text{ versus } H_{jk}^0: \trueimp_j \geq \trueimp_k$;
\item[(b)] 
A test of $H^1_{kj}:  \trueimp_k < \trueimp_j \text{ versus } H_{kj}^0: \trueimp_k \geq \trueimp_j.$
\end{enumerate}

Each test is composed of a p-value $p_{jk} = pairCompare(\basematrix_j, \basematrix_k)$
and a significance level $\alpha \in (0,0.5]$; the test rejects $H_{jk}^0$ if $p_{jk}<\alpha$. 
The test is \emph{calibrated} if: $\mathbb{P}(p_{jk} \leq \alpha) \leq \alpha \quad \text{for any } \trueimp_j \geq \trueimp_k$, meaning that the probability of rejecting $H^0_{jk}$ when $H^1_{jk}$ is correct is bounded by $\alpha$. For the tests to be calibrated, they need to account for the possible dependence between $\basematrix_j$ and $\basematrix_k$. In our implementation, we use the paired-sample t-test (see Section \ref{subsec:implementation}).

There is a natural relation between the results of the one-sided hypothesis tests and the ranking of the global FI values. The rejection of a hypothesis $H_{jk}^0: \trueimp_j \geq \trueimp_k$, implies acceptance of a \emph{partial ranking} of the global FI values, namely $\trueimp_j < \trueimp_k$. This partial ranking limits the global ranking: the feature $j$ cannot be ranked highest ($u_j < p$), and the feature $k$ cannot be ranked lowest ($l_k > 1$). We combine many partial ranking statements to improve the bounds on the ranks.

Combining many probabilistic decisions comes at a price. Setting the tests' rejection threshold to $\alpha$ limits the marginal probability of making an error in each partial ranking to $\alpha$. However, when combined, the probability of making at least one error increases with the number of tests, and without proper adjustments it may greatly exceed $\alpha$. In the next subsection, we adjust the p-values to control this probability over multiple tests.

\subsection{Controlling Partial Ranking Error}

We define $\paranking$ as the set of partial rankings from all pairwise tests 
$\paranking = \{(j,k) : H_{jk}^0 \text{ was rejected} \}$. A partial ranking error is a pair $(j',k')\in \paranking$ for which $\trueimp_{j'}\geq \trueimp_{k'}$. 
For simultaneous CIs, we want to control the probability of error over all the partial rankings.

\begin{definition}(Family-Wise Error Rate)
The family-wise error rate (FWER) is controlled at probability level $\alpha$ on the set of partial rankings $\paranking$ if the probability of making any partial ranking error is less than $\alpha$:
$\mathbb{P}(\exists (j',k') \in \paranking:  \trueimp_{j'} \geq \trueimp_{k'}) \leq \alpha$.
\end{definition}

To control the FWER, we replace the original p-values with a set of adjusted p-values $\mathbf{p}^{adj} = FWERAdjust(\basematrix, pairCompare)$. After adjustment, the partial rankings are obtained by comparing $p_{jk}^{adj}$ and $p_{kj}^{adj}$ to the required FWER level $\alpha \,.$\footnote{In practice, when both tests use the same data and the threshold $\alpha$ is less than $0.5$, none or just one of the null hypotheses will be rejected (there will not be a case in which both of the null hypotheses are rejected).} Some examples of adjustment procedures in which the FWER is controlled are provided in Section \ref{subsec:implementation}.

\subsection{Confident Simultaneous Feature Ranking}\label{subsec:alg}

When the FWER is controlled for the partial rankings set, we can use the partial rankings set to derive simultaneous CIs for the true ranks:

\begin{theorem} \citep{al2022simultaneous}\label{thm}
Let $\paranking$ be the set of partial rankings with FWER control at level $\alpha$. For $j = 1,...,p$, define: 
\begin{align*}
    L_j = 1 + \#\{k : (k,j) \in D\}, \\
    U_j = p - \#\{k : (j,k) \in D\}.
\end{align*}
Then the sets $\{[L_j, U_j]$ for $j \in \features\}$ are $(1-\alpha)$ simultaneous CIs for the true ranks.
\end{theorem}

The construction naturally extends the definition of rank-set provided in Definition \ref{def:setrank}. The idea of the proof is that a coverage failure means that the set of true (one-sided) differences is smaller than the set of observed (one-sided) differences. This means that at least one partial ranking in $\paranking$ is false. Therefore the FWER upper bounds the probability of an error in the CIs (see proof in Appendix \ref{app:proof}). Our ranking method is based on ICRanks \citep{al2022simultaneous}; the way in which the proposed method differs from ICRanks is discussed in Section \ref{subsec:implementation}.

Algorithm \ref{alg} summarizes our method for constructing simultaneous CIs for the true ranks. The algorithm works directly on the base FI matrix without requiring access to the trained model, the FI method, or the explanation set. The main assumption is that our paired test is calibrated for the possible distributions of base FI values.

\begin{algorithm}
\caption{Simultaneous CIs for Ranks}\label{alg}
\begin{algorithmic}
\REQUIRE 
\STATE $\basematrix$: base FI matrix;
\STATE $1 - \alpha > 0$: level of confidence;
\STATE $pairCompare$: suitable paired test;
\STATE $FWERAdjust$: 
FWER adjustment procedure.

\FOR {$j,k \in \features, j \neq k$} {
\STATE    $p_{jk} \gets pairCompare(\basematrix_j, \basematrix_k)$.
}\ENDFOR

\STATE $\mathbf{p}^{adj} =  \gets FWERAdjust(\basematrix, pairCompare)$
\STATE $D \gets  \{(j,k) : p_{jk}^{adj} \leq \alpha\}$

\FOR {$j \in \features$} {
\STATE $L_j \gets 1 + \#\{k : (k,j) \in \paranking\}$
\STATE $U_j \gets p - \#\{k : (j,k) \in \paranking\}$
}\ENDFOR

\RETURN $[L_1, U_1], \ldots, [L_p, U_p]$.
\end{algorithmic}
\end{algorithm}

\subsection{Ranking Method Implementation}\label{subsec:implementation}

\paragraph{Paired Test} We use a parametric paired t-test to compute p-values for the pairs of base FI values.
Set $\mathbf{d} = \basematrix_j - \basematrix_k$ to be the vector of differences, and 
denote $\bar{d}$ as the sample average and $s_d$ as the sample standard deviation. Then the one-sided $\alpha$ level test rejects the null hypothesis if  $\bar{d}/ (s_d/\sqrt{n}) >  T_{n-1}(1-\alpha)$, where $T_{n-1}(1-\alpha)$ denotes the $1-\alpha$ quantile of student-t ($n-1$ df). The paired t-test is fairly robust to departures from a normal distribution \citep{posten1979robustness}.

\paragraph{Adjustment for Multiple Tests} We implement two sequential procedures to adjust (increase) the p-values:
\begin{itemize}
    \item Holm's procedure \citep{holm1979simple}. Assuming that the base FI values are normally distributed, the paired t-test is calibrated with this procedure. We implement Holm's procedure on the one-sided hypothesis tests, although for approximately normal data, the two-sided Holm would likely work well also. See \cite{shaffer1980control, shaffer1995multiple} on using Holm's procedure for pairwise tests.
    \item Min-P \citep{westfall1993resampling}. This procedure is based on bootstrapping. Therefore, no further assumptions are required.
\end{itemize}
The adjusted p-values are then compared to the predefined threshold of $\alpha$ (details on the procedures are provided in Appendix \ref{app:fwer}). With these procedures, if the p-values are calibrated, then the FWER for the rejected tests is controlled at the $\alpha$ level, regardless of the dependence.

\paragraph{Comparison to ICRanks} Similar to Algorithm \ref{alg}, ICRanks \citep{al2022simultaneous} is based on Tukey's correction \citep{tukey1953problem} in order to control the differences between ranks simultaneously. Tukey's correction is designed for normal and independent data, which are distributional assumptions that would usually not hold for FI values. In contrast, our algorithm applies a test to each feature pair separately, and the Holm's or the Min-P procedure is performed on the resulting p-values; therefore, it can be utilized with robust or nonparametric tests \citep{wilcox2011introduction}.

\section{Evaluation}\label{sec:experiments}

\begin{figure}[ht]
     \centering
     \includegraphics[width=\columnwidth]{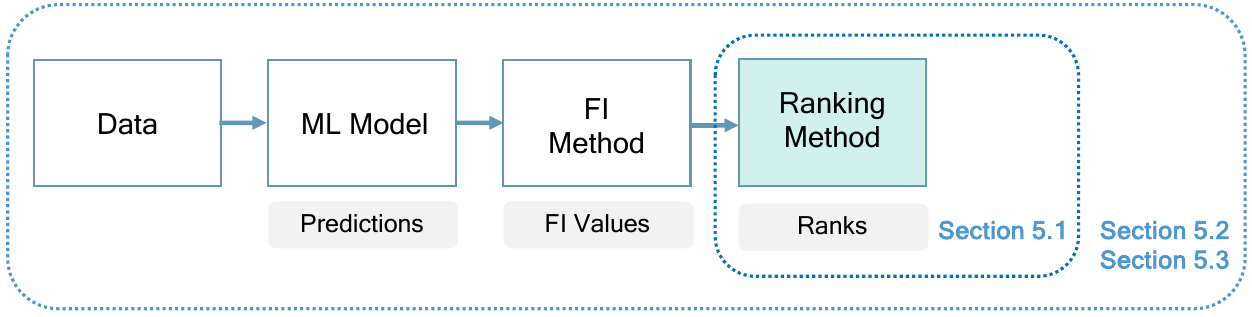}
     \caption[]%
     {Feature ranking and evaluation process.}
     \label{fig:process}
\end{figure}

In this section, we evaluate our base-to-global framework and ranking method. We use synthetic data to assess our method's validity (simultaneous coverage) and efficiency. We analyze our ranking method by generating base FI values directly (Section \ref{sec:comparison}). We note that feature ranking is an interpretability step at the end of an ML task, as shown in Figure \ref{fig:process}; therefore, we simulate the entire process of training and explaining a model with simulated data (Section \ref{sec:shap_simulated}) and real data (Section \ref{sec:real}).

\paragraph{Metrics} We use the metrics (ranking measures) suggested by \cite{al2022simultaneous} to define simultaneous coverage and efficiency:
\begin{itemize}
    \item \emph{Simultaneous coverage} -- the proportion of experiments where all true ranks are covered by their CIs: $one \text{ if all } \{\trueimp_j \in [L_j, U_j]\}, and zero \text{ otherwise} $.
    \item \emph{Efficiency} -- the average relative size of the CIs: $\frac{1}{p \cdot (p-1)}\sum_{j=1}^p (U_j - L_j)$.
\end{itemize}
Higher coverage and lower efficiency are better.

\subsection{Ranking Method Comparison}\label{sec:comparison}

\paragraph{Ranking Methods} We compare the ranking measures of four ranking methods: a naive ranking method based on empirical quantiles of bootstrap samples as a baseline (details are provided in Appendix \ref{app:naive}), ICRanks,\footnote{\href{https://cran.r-project.org/web/packages/ICRanks/ICRanks.pdf}{ICRanks package}} our ranking method with Holm's procedure, and our ranking method with the Min-P adjustment procedure.

We sample the base FI values from a multivariate-normal distribution $N_p(\mathbf{\mu}, \Sigma)$ with predetermined vector of means $\mathbf{\mu}$ and a covariance matrix $\Sigma$. The true global FI values are the means, and we control the correlation structure between the base FI values via the definition of the covariance matrix.

\paragraph{Vector of Means} The structure of the vector $\mathbf{\mu}$ is $\left(1^{\mu\text{-exponent}}, 2^{\mu\text{-exponent}}, \ldots, (p+1)^{\mu\text{-exponent}}\right)^T$, with $\mu$-exponent $\in [0.1, 0.25, 0.5]$. A lower value of $\mu$-exponent results in a more dense vector of means. Ties between the means are allowed.

\paragraph{Covariance Matrix} The covariance matrix structure is composed of a vector $\mathbf{\sigma}^2$ of the variances of the base FI values sampled from the re-scaled chi-squared distribution ($\chi_{(5)}^2 / 5$). The correlation matrix structure can be one of three structures: identity (no correlations), block-wise pairs, or equal correlations with $\rho \in [0.1, 0.5, 0.9]$. In addition, we vary the level of noise in the base FI values by scaling the vector $\mathbf{\sigma}$ by $\sigma$-factor $\in [0.2, 1, 5]$.

We analyze the ranking measures for multiple conditions of the vector of means ($\mathbf{\mu}$) and the correlation matrix ($\Sigma$) (a total of 486 conditions). The number of features $p$ is one of: $[10, 30, 50]$, and the number of base FI values $n$ is one of: $[100, 300, 1000]$. We sample 100 independent explanation sets for each configuration and report the average ranking measures across the repetitions. Below, we present the results for $p=30$, $\mu$-exponent=0.25, and equal correlations. Additional results are presented in Appendix \ref{app:mock}.

\paragraph{Simultaneous Coverage} In the naive ranking method, simultaneous coverage is not maintained in all conditions. All other methods maintain simultaneous coverage levels of almost 100\%; this indicates that they are overly conservative compared to the nominal required simultaneous coverage of $90\%$.

\begin{figure}[ht]
    \centering    
    \includegraphics[width=\columnwidth]{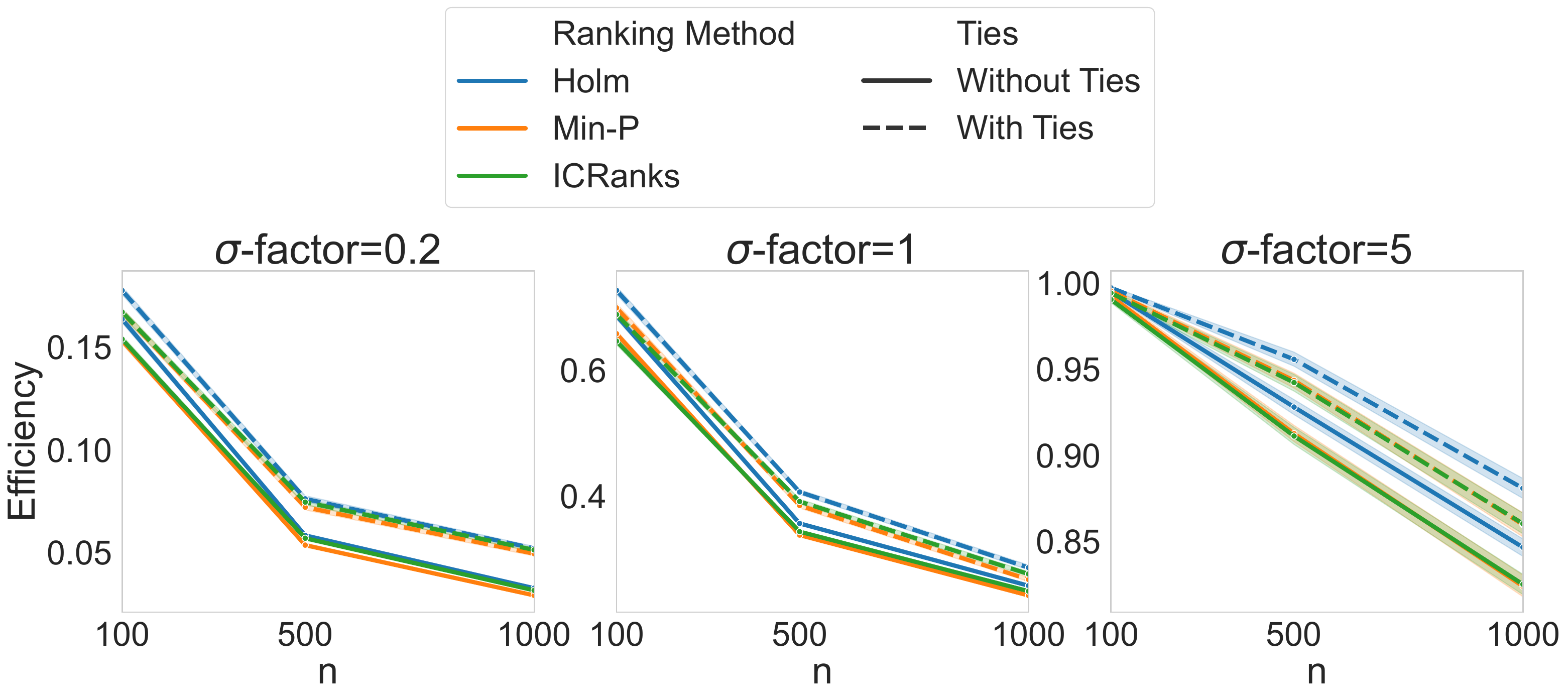}
    \caption[]%
    {Ranking efficiency as a function of $n$ for multiple $\sigma$-factors and three ranking methods. Low values mean smaller sets and are therefore better. The methods' efficiency is similar.}
    \label{fig:mock_exp_no_corr}
\end{figure}

\begin{figure}[ht]
     \centering
     \begin{subfigure}{\columnwidth}
     \centering
     \includegraphics[width=\columnwidth]{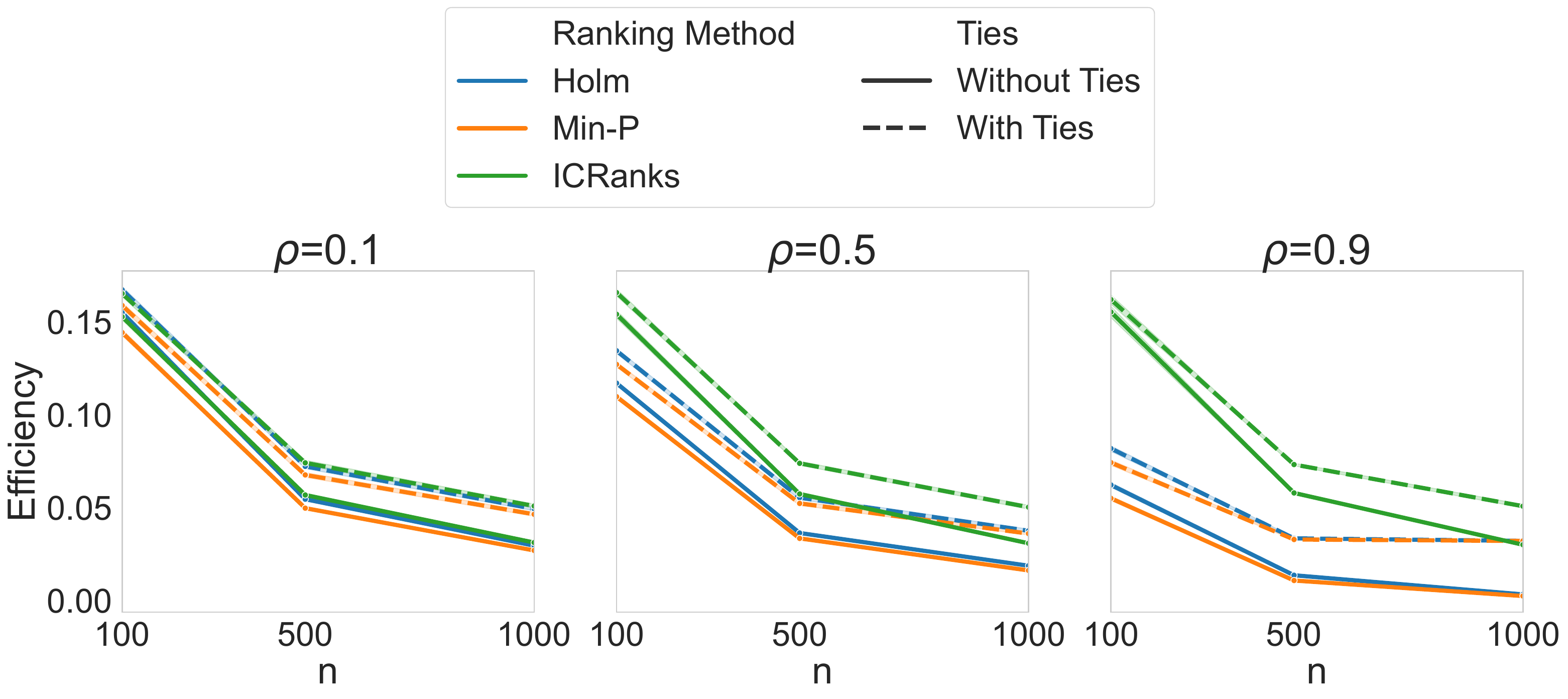}
     \caption{$\sigma$-factor=0.2}
     \end{subfigure}%
     \\
     \begin{subfigure}{\columnwidth}
     \centering
     \includegraphics[width=\columnwidth]{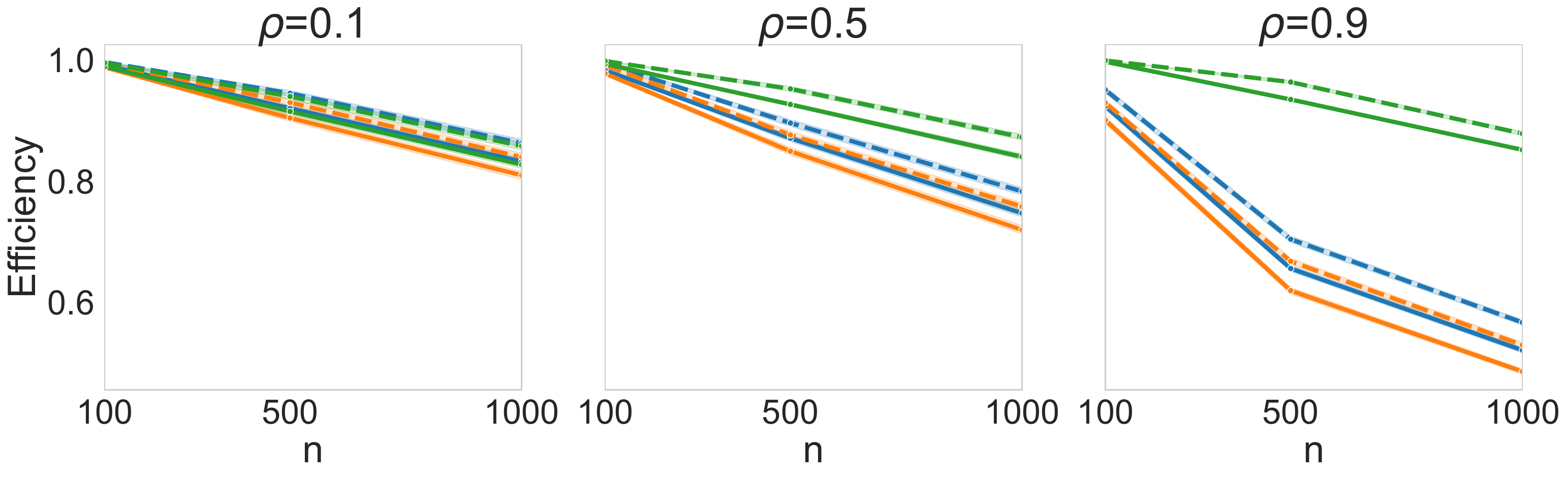}
     \caption{$\sigma$-factor=5}
     \end{subfigure}%
     \caption[]%
     {Ranking efficiency with low (a) and high (b) $\sigma$-factors, as a function of $n$ for multiple levels of correlations ($\rho$) and three ranking methods.}
     \label{fig:mock_exp_corr}
\end{figure}

\paragraph{Efficiency} Without correlations, efficiency degrades as the $\sigma$-factor increases, with almost no difference observed between the methods (Figure \ref{fig:mock_exp_no_corr}). With correlations, as $\rho$ increases our method becomes more efficient than ICRanks, with the Min-P adjustment seen to be slightly more efficient than Holm's procedure. The gap between the methods increases as the $\sigma$-factor increases (Figure \ref{fig:mock_exp_corr}).

\subsection{SHAP Ranking Measures}\label{sec:shap_simulated}

Here, we simulate the entire ML process as described in Figure \ref{fig:process} with simulated data. We analyze the ranking measures and runtime of our ranking method with Min-P and Holm's procedures compared to ICRanks.

\paragraph{Data Generating Process (DGP)} We follow the DGP of \cite{ishwaran2019standard}. We sample a data matrix $X$ with independent uniformly distributed features and calculate $Y$ as a function of $X$ with noise. We define two functions:
\begin{enumerate}[label=(\Alph*)]
    \item $ \! 
    \begin{aligned}[t]
    y &= 10\sin{(\pi x_1x_2)} + 20(x_3 - 0.5)^2 + 10x_4 + 5x_5 + \epsilon; \\
    & \{x_j\} \sim U(0, 1); \, \epsilon \sim N(0, 1).
\end{aligned}
$
\item $ \! 
\begin{aligned}[t]
    y &= (x_1^2 + [x_2x_3 - (x_2x_4)^{-1}]^2)^{0.5} + \epsilon; \\
    & x_1 \sim U(0, 10), x_2 \sim U(\pi, 2\pi), x_4 \sim U(1, 5), \\
    & \text{all other features } \{x_j\} \sim U(0, 1); \, \epsilon \sim N(0, 1).
\end{aligned}
$
\end{enumerate}

The definitions of $(X,Y)$ are for $p=10$ features. We simulate a larger number of features by defining the functions for cycles of 10 features.  For example, in DGP-A, $X_{11} \sim U(0, 10)$ and is added to $Y$ as $X_{11}^2$.

For this simulation, we sample a large data matrix $X_{M \times p}$ ($M=500k$), calculate $Y$ as a function of $X$ with noise, and train a prediction model on $\sampleset_{train}=(X, Y)$. We calculate the global FI values $\imp_1, \ldots \imp_p$ based on a sufficiently large sample ($n=1M$), making it a low variance estimator of $\trueimp_1, \ldots \trueimp_p$ \citep{slack2021reliable}. We generate multiple simulated datasets, varying the number of features ($p$) and base FI values ($n$), the DGP, and the prediction models. We sample 100 independent explanation sets for each evaluation configuration to measure the ranking efficiency, simultaneous coverage, and runtime.

Below we present the results for the DGP-A with a random forest (RF) model \citep{breiman2001random} and DGP-B with an XGBoost (XGB) model \citep{chen2016xgboost}  (see Appendix \ref{app:shap_simulated} for the complete results). We use TreeSHAP \citep{lundberg2019Tree} to compute the base FI values, relying on the definition of base and global FI values presented in Section \ref{sec:framework} (Equation \ref{eq:local_shap}). To calculate the ranking measures, we repeatedly sample $\sampleset_{explain}$ independent of $\sampleset_{train}$.

\paragraph{Simultaneous Coverage} All of the examined methods maintain simultaneous coverage levels of almost 100\% in all simulated conditions. However when the base FI values have an extremely long tail, simultaneous coverage is not guaranteed (an example is provided in Appendix \ref{app:shap_simulated_non_normal}).

\begin{figure}[ht]
     \centering
     \begin{subfigure}{\columnwidth}
     \centering
     \includegraphics[width=\columnwidth]{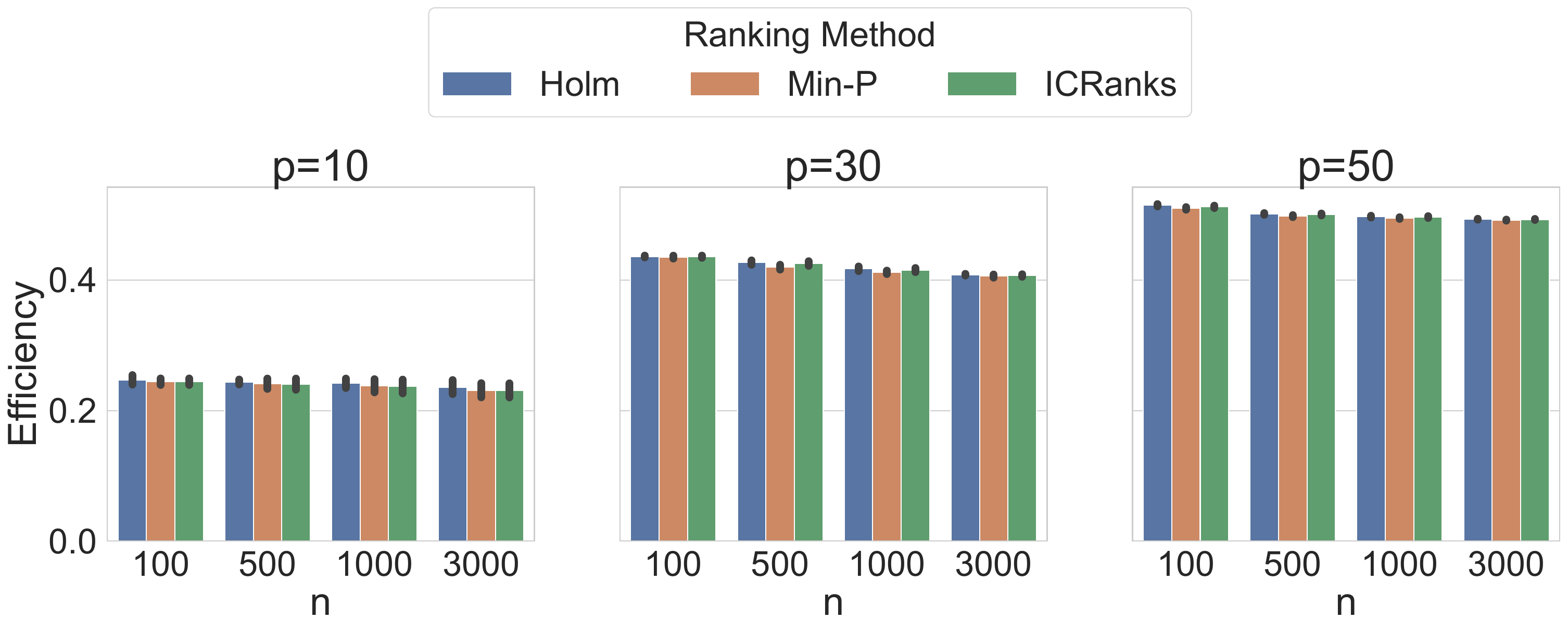}
     \caption{RF example with DGP-A}
     \end{subfigure}%
     \\
     \begin{subfigure}{\columnwidth}
     \centering
     \includegraphics[width=\columnwidth]{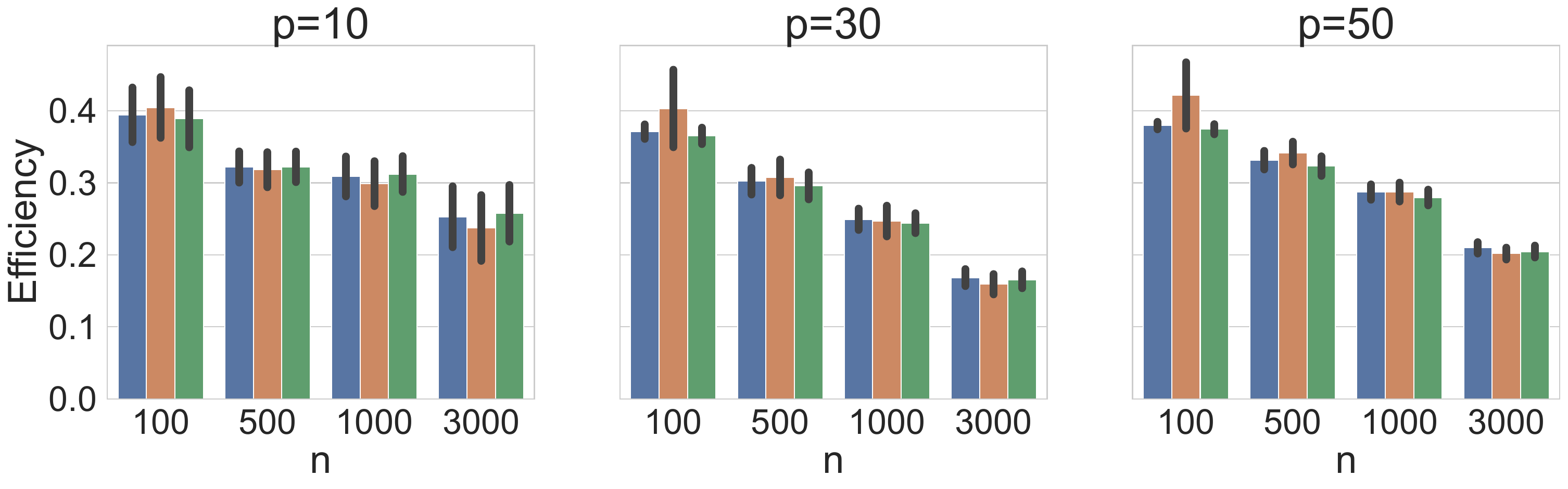}
     \caption{XGB example with DGP-B}
     \end{subfigure}%
     \caption[]%
     {Ranking efficiency as a function of $n$ for different numbers of features ($p$) and ranking methods. The efficiency improves as $n$ increases.}
     \label{fig:shap_results}
\end{figure}

\paragraph{Efficiency} We can see that ICRanks is comparable to our ranking method. The ranking efficiency improves as $n$ increases (Figure \ref{fig:shap_results}). For the XGB model, we see that the efficiency of our method with the Min-P procedure is worse than that of our method with Holm's procedure for low $n$ values; the Min-P procedure recalibrates the p-values based on resampled data, which is an inefficient process when $n$ is low.

\paragraph{Runtime} We analyze the runtime of TreeSHAP (computation of base FI values) and the ranking times (ICRanks, Holm's procedure, and the Min-P procedure). The runtime of ICRanks and Holm's procedure is ten times faster than the runtime of TreeSHAP. The Min-P procedure requires B repetitions (bootstrap samples) of the pairwise tests, so the runtime increases with B. Details are provided in Appendix \ref{app:shap_simulated_runtime}.

\subsection{Real Data Experiments}\label{sec:real}

\subsubsection{Ranking Stability}

\begin{figure}[ht]
     \centering
     \includegraphics[width=\columnwidth]{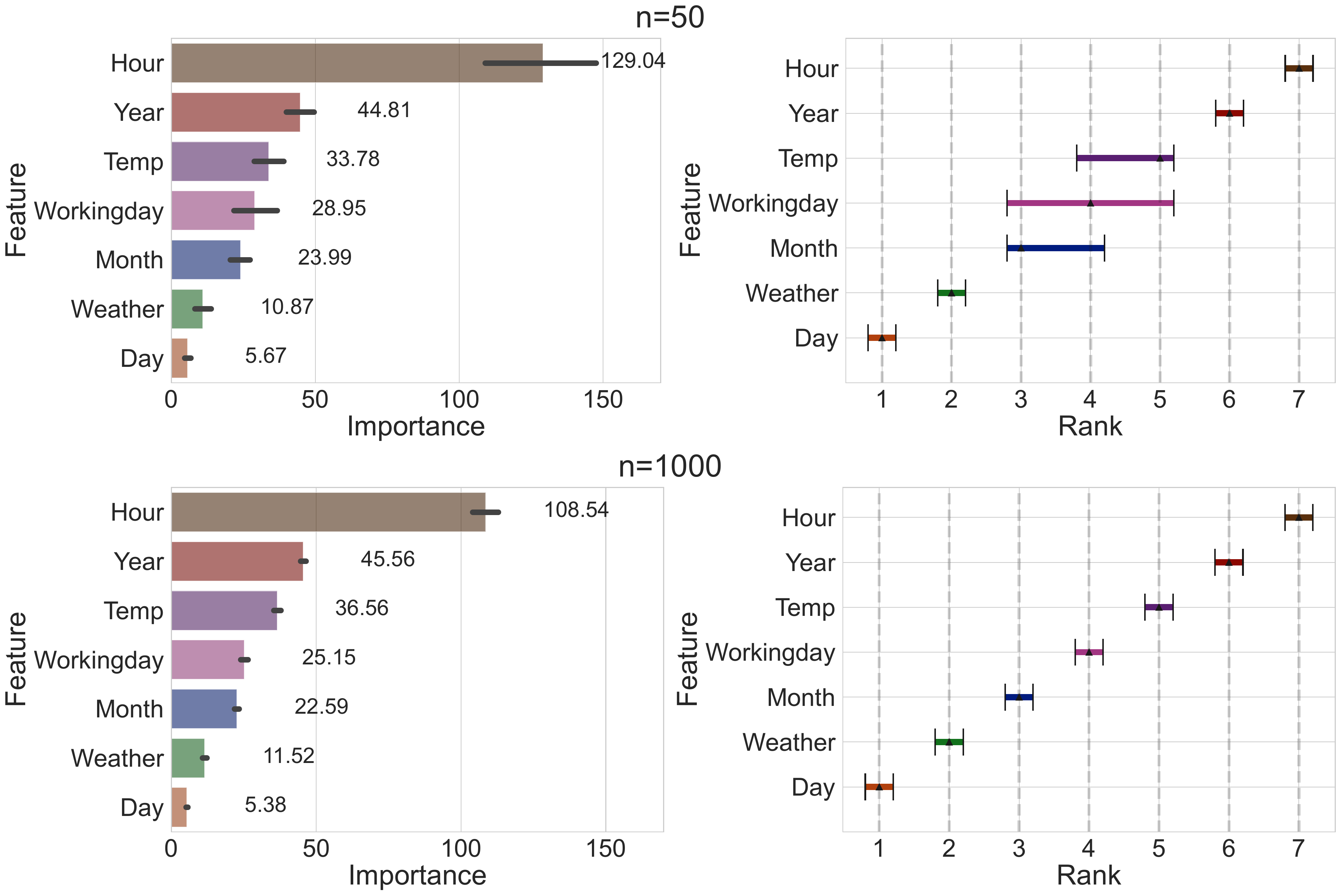}
     \caption[]%
     {Global SHAP values (left) and CIs for the true ranks (right) for the bike sharing dataset. The importance values were obtained from 50 (top) and 1,000 (bottom) observations. The CIs point out uncertain feature rankings for a small sample size.}
     \label{fig:bike_base_size}
\end{figure}

We demonstrate the use of our ranking method and present the simultaneous CIs produced with the bike sharing dataset \citep{fanaee2014event}, the TreeSHAP FI method, and Holm's procedure. We create 60/40 train/test splits, fit an XGB regression model (default hyperparameters) to the training set ($R^2=0.98$), and evaluate the performance on the test set ($R^2=0.94$). Then, we calculate the base FI values for $n=50$ and $n=1000$ by sampling from the test set. Presenting the CIs for the ranks enables us to compare the stability for different sizes of $n$ (see Figure \ref{fig:bike_base_size}). The triangles within the CIs are the observed global FI values. The process of constructing the CIs for $n=50$ base FI values is described in Appendix \ref{app:full_example}.

\subsubsection{Training Stability}

\begin{figure}[ht]
     \centering
     \includegraphics[width=\columnwidth]{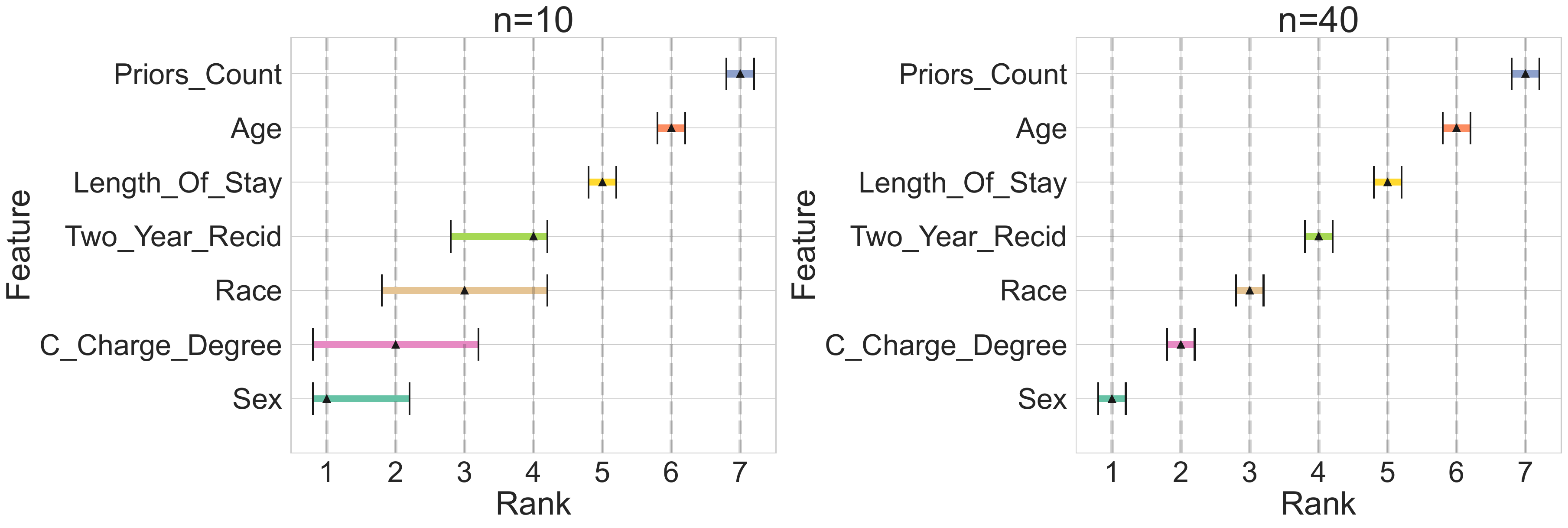}
     \caption[]%
     {CIs for the true ranks for $n=10$ (left) and $n=40$ (right) trained models. The CIs present uncertainty in training a model based on $M=3K$ training observations.}
     \label{fig:compas_training}
\end{figure}

Our base-to-global framework can also be used to quantify the uncertainty in training stemming from the sampling of the training set. Here, we use the COMPAS dataset \citep{angwin2016machine}, an RF classification model (default hyperparameters), the PFI method, and the Min-P procedure. We define the base FI values as global PFI values. Each trained model produces a base FI vector $\basevalue_i$; the global FI values are obtained by resampling and training multiple equivalent models (with the same hyperparameters and size of $\sampleset_{train}$ ($M=3K$) and similar training accuracy ($0.883 \pm 0.005$)). We use the same explanation set ($N=600$) to calculate the importance values. Figure \ref{fig:compas_training} presents the true ranks' CIs for two values of $n$. The observed uncertainty in the ranking for $n=10$ indicates that the randomness in sampling can influence the learned mapping  between the features and the target variable.

\subsubsection{High-Dimensional Data}

\begin{figure}[ht]
     \centering
     \includegraphics[width=0.45\columnwidth]{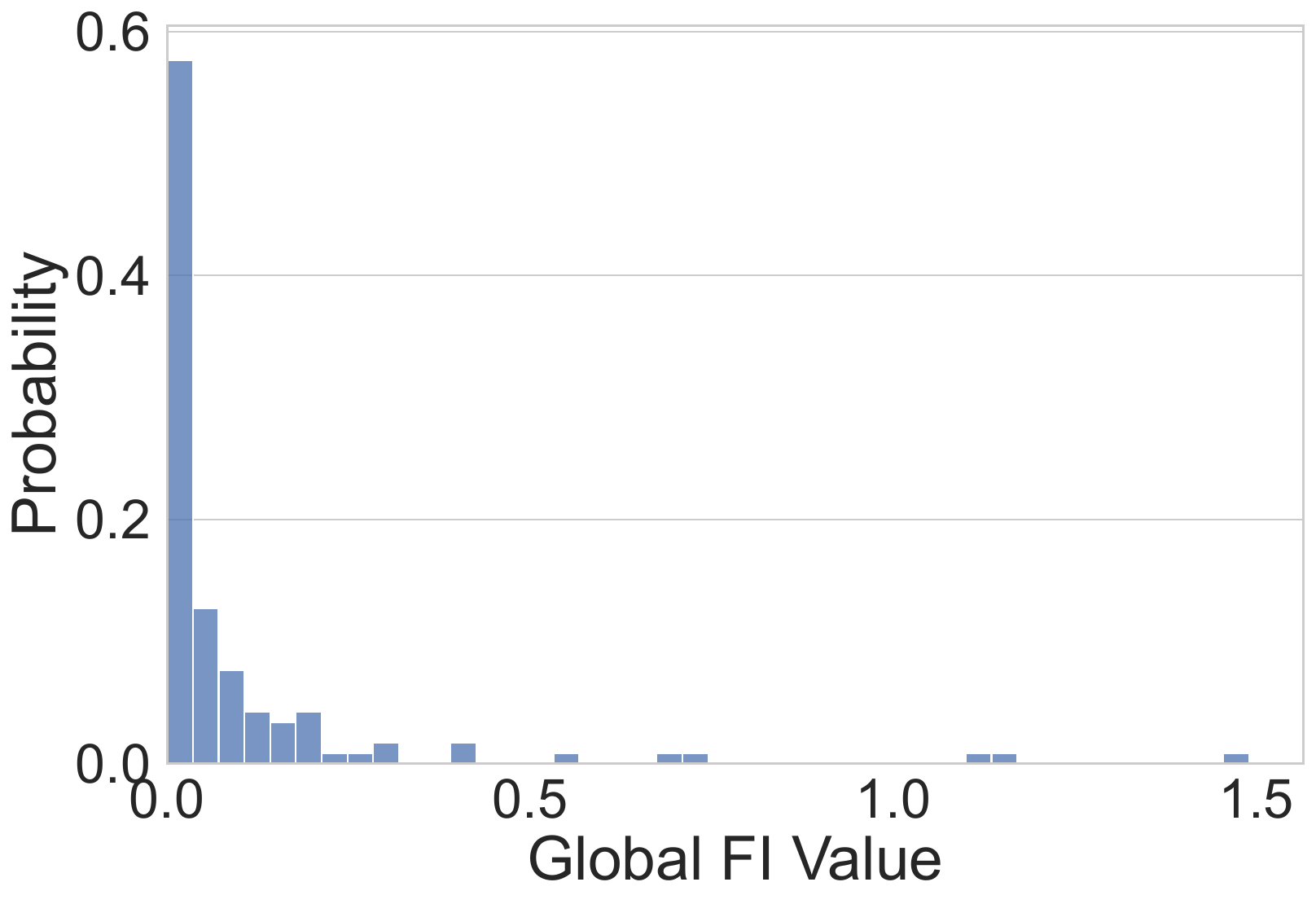}
     \includegraphics[width=0.45\columnwidth]{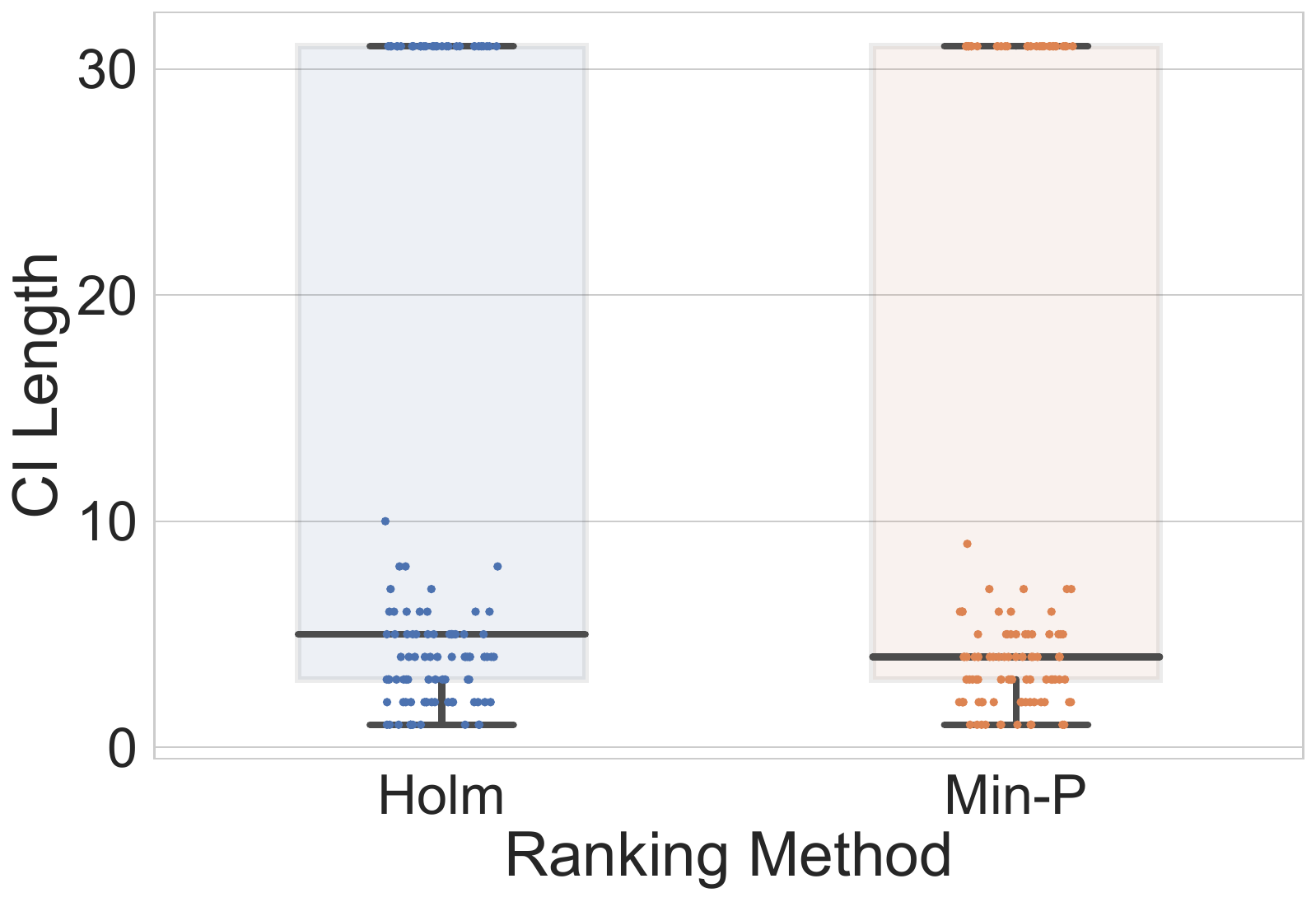}
     \caption[]%
     {Global SHAP FI value distribution (left) and length of the CIs for the ranks (right) for the Nomao dataset features. The FI value of many features is zero, and the ranking is accordingly inefficient.}
     \label{fig:big_data}
\end{figure}

In previous sections (Sections \ref{sec:comparison} and \ref{sec:shap_simulated}), we analyzed the validity of our ranking method in multiple settings, including with moderately high-dimensional data ($p=50$), and showed that our method maintains simultaneous coverage. Now we demonstrate the use of our ranking method with high-dimensional data, utilizing the Nomao dataset \citep{misc_nomao_227}, which consists of 118 input features and a binary target variable. We create 60/40 train/test splits, fit an XGB classification model (with the default hyperparameters) to the training set ($accuracy=0.99$), and evaluate the performance on the test set ($accuracy=0.97$). Then, we calculate the base FI values with TreeSHAP; the distribution of the global FI values is shown on the left side of Figure \ref{fig:big_data}. Thirty-one features have importance values of zero.

We use our ranking method to rank all the features from the least important (1) to the most important (118); the full ranking is presented in Appendix \ref{app:high-dim} (Figure \ref{fig:high_dim_ranking}). As 31 features have the same importance value of zero, the CI of each feature is $[1, 31]$. In such a case, if we measure the efficiency of the ranking across all features, the long CIs of the irrelevant features will affect it. The skewness of the CI length is presented on the right of Figure \ref{fig:big_data}. A model will likely use some features, and the unused features will get a low importance value (or a value of zero); a filtering step is required to improve the ranking process by comparing only the relevant features.

\section{Related Work}\label{sec:related}

\subsection{Uncertainty in Feature Ranking}
Ordering features in terms of their importance to the model's prediction is referred to as feature ranking. It is often infeasible to determine the ``right" order or perfect subset of features, because it requires the examination of all possible feature subsets \citep{prati2012combining}. Many studies suggested overcoming this limitation and obtaining a more stable ranking by applying a two-step procedure in which multiple rankings are generated and the outputs are combined into a single ranking \citep{saeys2008robust}. Each of the rankings generated in the first step is obtained by ordering the global FI values, which are produced using different FI methods \citep{schulz2021uncertainty} or by resampling the data and ranking the output FI values using a single FI method \citep{vettoretti2021variable, alaiz2020information, salman2022stability}. The process of combining all of the rankings into a single ranking is sometimes based on voting \citep{vettoretti2021variable, schulz2021uncertainty}, pairwise comparisons of the rankings \citep{prati2012combining, salman2022stability}, or other techniques \citep{alaiz2020information}. In all of the techniques mentioned above, more stable ranking is achieved by aggregating multiple global scores or rankings, a process that is computationally expensive and requires many explanation sets or FI methods. In contrast, our ranking method produces a stable ranking based on a single FI method and explanation set.

\subsection{Ranking and Selection}
The problem of ranking and selection (R \& S) of items has been well studied by researchers in the field of statistics, and various solutions have been suggested \citep{gupta1965some, boesel2003using}. Some studies focused on ranking items based on noisy data and pairwise comparisons \citep{wright2014ranking, valdeira2022ranking}. Other studies proposed methods that look for the best item \citep{eckman2020revisiting}, select the set of top- or lowest-ranked items, or, most similar to our work, methods that rank all items based on the observed means \citep{zhang2014confidence, klein2020joint, wright2014ranking}. After ranking all of the items, a subset of items might be selected.  \citep{al2022simultaneous, rising2021uncertainty}. Other researchers have proposed methods like ours that deal with the effect of multiple tests and examine how to control the FWER and increase the probability that the correct items are selected \citep{garcia2008extension, holm2013confidence}.

\section{Conclusions}

We propose a base-to-global framework and a method for constructing CIs for the true ranks of the global FI values. Because rankings are frequently used to summarize FI methods' output, it is crucial to consider the rankings' stability. Our method can be used with robust and nonparametric paired-tests to support non-standard FI distributions.

We present a rigorous criterion  for quantifying uncertainty that can be explicitly modeled (e.g., the explanation set size). We view the proposed method as a step toward producing new forms of stability assessments for explainable ML. In future research, we aim to address other sources of instability, such as the difference between FI methods, although their effect is more challenging to quantify. 

Finally, our current algorithm is conservative, as demonstrated in simulations where the coverage level surpasses the requested (1-$\alpha$)\%. Future research will also be aimed at narrowing the CIs while maintaining nominal coverage and reducing the impact of the number of features on coverage by using a filtering step (e.g., eliminating non-important features).

\subsubsection*{Acknowledgements}
We thank Noga H. Rotman for valuable discussions and advice. This work was supported by the Israel Science Foundation and the Center for Interdisciplinary Research at Hebrew University.

\newpage
\medskip
{
\small
\bibliography{ranking_arxiv}

\begin{thebibliography}{10}

\bibitem{bhardwaj2017study}
Rohan Bhardwaj, Ankita~R Nambiar, and Debojyoti Dutta.
\newblock A study of machine learning in healthcare.
\newblock In {\em 2017 IEEE 41st annual computer software and applications conference (COMPSAC)}, volume~2, pages 236--241. IEEE, 2017.

\bibitem{rundo2019machine}
Francesco Rundo, Francesca Trenta, Agatino~Luigi di~Stallo, and Sebastiano Battiato.
\newblock Machine learning for quantitative finance applications: A survey.
\newblock {\em Applied Sciences}, 9(24):5574, 2019.

\bibitem{deiana2022applications}
Allison~McCarn Deiana, Nhan Tran, Joshua Agar, Michaela Blott, Giuseppe Di~Guglielmo, Javier Duarte, Philip Harris, Scott Hauck, Mia Liu, Mark~S Neubauer, et~al.
\newblock Applications and techniques for fast machine learning in science.
\newblock {\em Frontiers in big Data}, 5:787421, 2022.

\bibitem{li2022machine}
Zhanzhao Li, Jinyoung Yoon, Rui Zhang, Farshad Rajabipour, Wil~V Srubar~III, Ismaila Dabo, and Aleksandra Radli{\'n}ska.
\newblock Machine learning in concrete science: applications, challenges, and best practices.
\newblock {\em npj Computational Materials}, 8(1):127, 2022.

\bibitem{preece2018stakeholders}
Alun Preece, Dan Harborne, Dave Braines, Richard Tomsett, and Supriyo Chakraborty.
\newblock Stakeholders in explainable ai.
\newblock {\em arXiv preprint arXiv:1810.00184}, 2018.

\bibitem{goodman2017european}
Bryce Goodman and Seth Flaxman.
\newblock European union regulations on algorithmic decision-making and a “right to explanation”.
\newblock {\em AI magazine}, 38(3):50--57, 2017.

\bibitem{breiman2001random}
Leo Breiman.
\newblock Random forests.
\newblock {\em Machine learning}, 45(1):5--32, 2001.

\bibitem{lundberg2017SHAP}
Scott Lundberg and Su-In Lee.
\newblock A unified approach to interpreting model predictions.
\newblock {\em arXiv preprint arXiv:1705.07874}, 2017.

\bibitem{lundberg2019Tree}
Scott Lundberg, Gabriel Erion, Hugh Chen, Alex DeGrave, Jordan~M Prutkin, Bala Nair, Ronit Katz, Jonathan Himmelfarb, Nisha Bansal, and Su-In Lee.
\newblock Explainable ai for trees: From local explanations to global understanding.
\newblock {\em arXiv preprint arXiv:1905.04610}, 2019.

\bibitem{merrick2020explanation}
Luke Merrick and Ankur Taly.
\newblock The explanation game: Explaining machine learning models using shapley values.
\newblock In {\em International Cross-Domain Conference for Machine Learning and Knowledge Extraction}, pages 17--38. Springer, 2020.

\bibitem{covert2020imputation}
Ian Covert, Scott Lundberg, and Su-In Lee.
\newblock Explaining by removing: A unified framework for model explanation.
\newblock {\em arXiv preprint arXiv:2011.14878}, 2020.

\bibitem{molnar2020general}
Christoph Molnar, Gunnar K{\"o}nig, Julia Herbinger, Timo Freiesleben, Susanne Dandl, Christian~A Scholbeck, Giuseppe Casalicchio, Moritz Grosse-Wentrup, and Bernd Bischl.
\newblock General pitfalls of model-agnostic interpretation methods for machine learning models.
\newblock In {\em International Workshop on Extending Explainable AI Beyond Deep Models and Classifiers}, pages 39--68. Springer, 2020.

\bibitem{marx2023but}
Charles Marx, Youngsuk Park, Hilaf Hasson, Yuyang Wang, Stefano Ermon, and Luke Huan.
\newblock But are you sure? an uncertainty-aware perspective on explainable ai.
\newblock In {\em International Conference on Artificial Intelligence and Statistics}, pages 7375--7391. PMLR, 2023.

\bibitem{lakkaraju2020robust}
Himabindu Lakkaraju, Nino Arsov, and Osbert Bastani.
\newblock Robust and stable black box explanations.
\newblock In {\em International Conference on Machine Learning}, pages 5628--5638. PMLR, 2020.

\bibitem{agarwal2022rethinking}
Chirag Agarwal, Nari Johnson, Martin Pawelczyk, Satyapriya Krishna, Eshika Saxena, Marinka Zitnik, and Himabindu Lakkaraju.
\newblock Rethinking stability for attribution-based explanations.
\newblock {\em arXiv preprint arXiv:2203.06877}, 2022.

\bibitem{slack2021reliable}
Dylan Slack, Anna Hilgard, Sameer Singh, and Himabindu Lakkaraju.
\newblock Reliable post hoc explanations: Modeling uncertainty in explainability.
\newblock {\em Advances in Neural Information Processing Systems}, 34:9391--9404, 2021.

\bibitem{ahn2023local}
Surin Ahn, Justin Grana, Yafet Tamene, and Kristian Holsheimer.
\newblock Local model explanations and uncertainty without model access.
\newblock {\em arXiv preprint arXiv:2301.05761}, 2023.

\bibitem{ishwaran2019standard}
Hemant Ishwaran and Min Lu.
\newblock Standard errors and confidence intervals for variable importance in random forest regression, classification, and survival.
\newblock {\em Statistics in medicine}, 38(4):558--582, 2019.

\bibitem{covert2020SAGE}
Ian Covert, Scott Lundberg, and Su-In Lee.
\newblock Understanding global feature contributions through additive importance measures.
\newblock {\em arXiv preprint arXiv:2004.00668}, 2020.

\bibitem{molnar2021relating}
Christoph Molnar, Timo Freiesleben, Gunnar K{\"o}nig, Giuseppe Casalicchio, Marvin~N Wright, and Bernd Bischl.
\newblock Relating the partial dependence plot and permutation feature importance to the data generating process.
\newblock {\em arXiv preprint arXiv:2109.01433}, 2021.

\bibitem{jaxa2021sources}
Marc Jaxa-Rozen and Evelina Trutnevyte.
\newblock Sources of uncertainty in long-term global scenarios of solar photovoltaic technology.
\newblock {\em Nature Climate Change}, 11(3):266--273, 2021.

\bibitem{heldt2021early}
Frank~S Heldt, Marcela~P Vizcaychipi, Sophie Peacock, Mattia Cinelli, Lachlan McLachlan, Fernando Andreotti, Stojan Jovanovi{\'c}, Robert D{\"u}richen, Nadezda Lipunova, Robert~A Fletcher, et~al.
\newblock Early risk assessment for covid-19 patients from emergency department data using machine learning.
\newblock {\em Scientific reports}, 11(1):4200, 2021.

\bibitem{rising2021uncertainty}
Justin Rising.
\newblock Uncertainty in ranking.
\newblock {\em arXiv preprint arXiv:2107.03459}, 2021.

\bibitem{al2022simultaneous}
Diaa Al~Mohamad, Jelle~J Goeman, and Erik~W van Zwet.
\newblock Simultaneous confidence intervals for ranks with application to ranking institutions.
\newblock {\em Biometrics}, 78(1):238--247, 2022.

\bibitem{benjamini2005false}
Yoav Benjamini and Daniel Yekutieli.
\newblock False discovery rate--adjusted multiple confidence intervals for selected parameters.
\newblock {\em Journal of the American Statistical Association}, 100(469):71--81, 2005.

\bibitem{hsu1996multiple}
Jason Hsu.
\newblock {\em Multiple comparisons: theory and methods}.
\newblock CRC Press, 1996.

\bibitem{ein2006thousands}
Liat Ein-Dor, Or~Zuk, and Eytan Domany.
\newblock Thousands of samples are needed to generate a robust gene list for predicting outcome in cancer.
\newblock {\em Proceedings of the National Academy of Sciences}, 103(15):5923--5928, 2006.

\bibitem{posten1979robustness}
Harry~O Posten.
\newblock The robustness of the one-sample t-test over the pearson system.
\newblock {\em Journal of Statistical Computation and Simulation}, 9(2):133--149, 1979.

\bibitem{holm1979simple}
Sture Holm.
\newblock A simple sequentially rejective multiple test procedure.
\newblock {\em Scandinavian journal of statistics}, pages 65--70, 1979.

\bibitem{shaffer1980control}
Juliet~Popper Shaffer.
\newblock Control of directional errors with stagewise multiple test procedures.
\newblock {\em The Annals of Statistics}, 8(6):1342--1347, 1980.

\bibitem{shaffer1995multiple}
Juliet~Popper Shaffer.
\newblock Multiple hypothesis testing.
\newblock {\em Annual review of psychology}, 46(1):561--584, 1995.

\bibitem{westfall1993resampling}
Peter~H Westfall and S~Stanley Young.
\newblock {\em Resampling-based multiple testing: Examples and methods for p-value adjustment}, volume 279.
\newblock John Wiley \& Sons, 1993.

\bibitem{tukey1953problem}
John~Wilder Tukey.
\newblock The problem of multiple comparisons.
\newblock {\em Multiple comparisons}, 1953.

\bibitem{wilcox2011introduction}
Rand~R Wilcox.
\newblock {\em Introduction to robust estimation and hypothesis testing}.
\newblock Academic press, 2011.

\bibitem{chen2016xgboost}
Tianqi Chen and Carlos Guestrin.
\newblock Xgboost: A scalable tree boosting system.
\newblock In {\em Proceedings of the 22nd acm sigkdd international conference on knowledge discovery and data mining}, pages 785--794, 2016.

\bibitem{fanaee2014event}
Hadi Fanaee-T and Joao Gama.
\newblock Event labeling combining ensemble detectors and background knowledge.
\newblock {\em Progress in Artificial Intelligence}, 2:113--127, 2014.

\bibitem{angwin2016machine}
Julia Angwin, Jeff Larson, Surya Mattu, and Lauren Kirchner.
\newblock Machine bias. propublica, may 23, 2016, 2016.

\bibitem{misc_nomao_227}
Laurent Candillier and Vincent Lemaire.
\newblock {Nomao}.
\newblock UCI Machine Learning Repository, 2012.
\newblock {DOI}: https://doi.org/10.24432/C53G79.

\bibitem{prati2012combining}
Ronaldo~C Prati.
\newblock Combining feature ranking algorithms through rank aggregation.
\newblock In {\em The 2012 international joint conference on neural networks (IJCNN)}, pages 1--8. Ieee, 2012.

\bibitem{saeys2008robust}
Yvan Saeys, Thomas Abeel, and Yves Van~de Peer.
\newblock Robust feature selection using ensemble feature selection techniques.
\newblock In {\em Machine Learning and Knowledge Discovery in Databases: European Conference, ECML PKDD 2008, Antwerp, Belgium, September 15-19, 2008, Proceedings, Part II 19}, pages 313--325. Springer, 2008.

\bibitem{schulz2021uncertainty}
Jonas Schulz, Rafael Poyiadzi, and Raul Santos-Rodriguez.
\newblock Uncertainty quantification of surrogate explanations: An ordinal consensus approach.
\newblock {\em arXiv preprint arXiv:2111.09121}, 2021.

\bibitem{vettoretti2021variable}
Martina Vettoretti and Barbara Di~Camillo.
\newblock A variable ranking method for machine learning models with correlated features: in-silico validation and application for diabetes prediction.
\newblock {\em Applied Sciences}, 11(16):7740, 2021.

\bibitem{alaiz2020information}
Rocio Alaiz-Rodriguez and Andrew~C Parnell.
\newblock An information theoretic approach to quantify the stability of feature selection and ranking algorithms.
\newblock {\em Knowledge-Based Systems}, 195:105745, 2020.

\bibitem{salman2022stability}
Reem Salman, Ayman Alzaatreh, and Hana Sulieman.
\newblock The stability of different aggregation techniques in ensemble feature selection.
\newblock {\em Journal of Big Data}, 9(1):1--23, 2022.

\bibitem{gupta1965some}
Shanti~S Gupta.
\newblock On some multiple decision (selection and ranking) rules.
\newblock {\em Technometrics}, 7(2):225--245, 1965.

\bibitem{boesel2003using}
Justin Boesel, Barry~L Nelson, and Seong-Hee Kim.
\newblock Using ranking and selection to “clean up” after simulation optimization.
\newblock {\em Operations Research}, 51(5):814--825, 2003.

\bibitem{wright2014ranking}
Tommy Wright, Martin Klein, and Jerzy Wieczorek.
\newblock Ranking populations based on sample survey data.
\newblock {\em Statistics}, page~12, 2014.

\bibitem{valdeira2022ranking}
Filipa Valdeira and Cl{\'a}udia Soares.
\newblock Ranking with confidence for large scale comparison data.
\newblock {\em arXiv preprint arXiv:2202.01670}, 2022.

\bibitem{eckman2020revisiting}
David~J Eckman, Matthew Plumlee, and Barry~L Nelson.
\newblock Revisiting subset selection.
\newblock In {\em 2020 Winter Simulation Conference (WSC)}, pages 2972--2983. IEEE, 2020.

\bibitem{zhang2014confidence}
Shunpu Zhang, Jun Luo, Li~Zhu, David~G Stinchcomb, Dave Campbell, Ginger Carter, Scott Gilkeson, and Eric~J Feuer.
\newblock Confidence intervals for ranks of age-adjusted rates across states or counties.
\newblock {\em Statistics in Medicine}, 33(11):1853--1866, 2014.

\bibitem{klein2020joint}
Martin Klein, Tommy Wright, and Jerzy Wieczorek.
\newblock A joint confidence region for an overall ranking of populations.
\newblock {\em Journal of the Royal Statistical Society: Series C (Applied Statistics)}, 69(3):589--606, 2020.

\bibitem{garcia2008extension}
Salvador Garcia and Francisco Herrera.
\newblock An extension on" statistical comparisons of classifiers over multiple data sets" for all pairwise comparisons.
\newblock {\em Journal of machine learning research}, 9(12), 2008.

\bibitem{holm2013confidence}
Sture Holm.
\newblock {\em Confidence intervals for ranks}.
\newblock 2013.

\bibitem{efron2012large}
Bradley Efron.
\newblock {\em Large-scale inference: empirical Bayes methods for estimation, testing, and prediction}, volume~1.
\newblock Cambridge University Press, 2012.

\end{thebibliography}
}

\newpage
\section*{Appendix}
\appendix
\section{PFI Variance Analysis}\label{app:pfi_var}

In Section \ref{sec:framework}, we present two options for defining the base FI values for PFI \citep{breiman2001random}: (1) a single permutation, and (2) a single observation. We obtain the same global FI values from the two base definitions by setting the values of the number of permutations ($B$) and the size of the explanation set ($N$) accordingly. However, the different decomposition of the global FI value to base FI values allows for the analysis of various sources of uncertainty -- the variance of the permutations and the variance of the explanation set. Our framework is limited to quantifying only one source of uncertainty by aggregating base FI values to global FI values. Therefore, using it might be a problem when the uncertainty of the global FI values stems from multiple sources of uncertainty. Nevertheless, if most of the variability comes from one of the sources, it is reasonable to target it and disregard the other sources. In the case of PFI, we expect that the size of the explanation set introduces greater variance than the number of permutations. Our results clearly show this; therefore, we can use our framework to quantify the uncertainty of global PFI values.

\subsection{Experimental Setup}

In this experiment, we use the same DGPs (A and B) described in Section \ref{sec:shap_simulated}, including the definition of $X$, $Y$, and the functions.

\subsubsection{Dummy Prediction Model} 

Instead of training a model, we create a \emph{Dummy} model that predicts $Y$ from $X$ using the DGP's function. We use this approach to control the variability stemming from the training and focus on the variability of the permutations and explanation set.

\subsubsection{Experiment Details}

We sample the data as described above with various configurations of $B$, $p$, and $N$, and the two functions. For each configuration:
\begin{enumerate}[label=(\arabic*)]
    \item We perform $B$ permutations for each observation and calculate the loss difference for each permutation $b$: $L(f(x_{[j]}^b), y) - L(f(x), y)$.
    \item We average all of the permutations for each observation.
    \item We average all of the observations.
\end{enumerate}
The result of steps 1-3 is a set of $p$ global FI values $\imp_1^{PFI}, \ldots, \imp_p^{PFI}$. We repeat this process 100 times and calculate the average and standard deviation (SD) across the repetitions.

\subsection{Results}

\begin{figure}[ht]
     \centering
     \begin{subfigure}{\columnwidth}
     \centering     \includegraphics[width=\columnwidth]{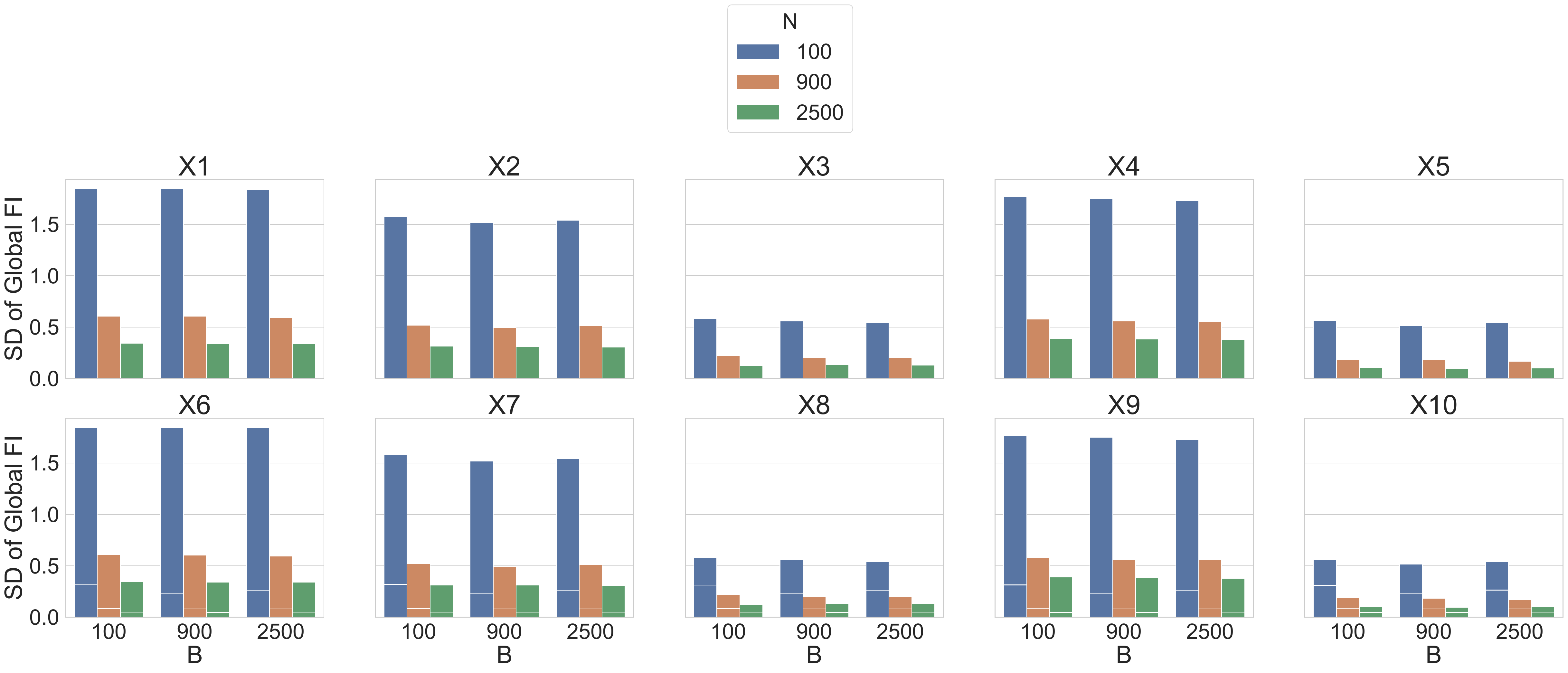}
     \caption{DGP-A}
     \end{subfigure}%
     \\
     \begin{subfigure}{\columnwidth}
     \centering
     \includegraphics[width=\columnwidth]{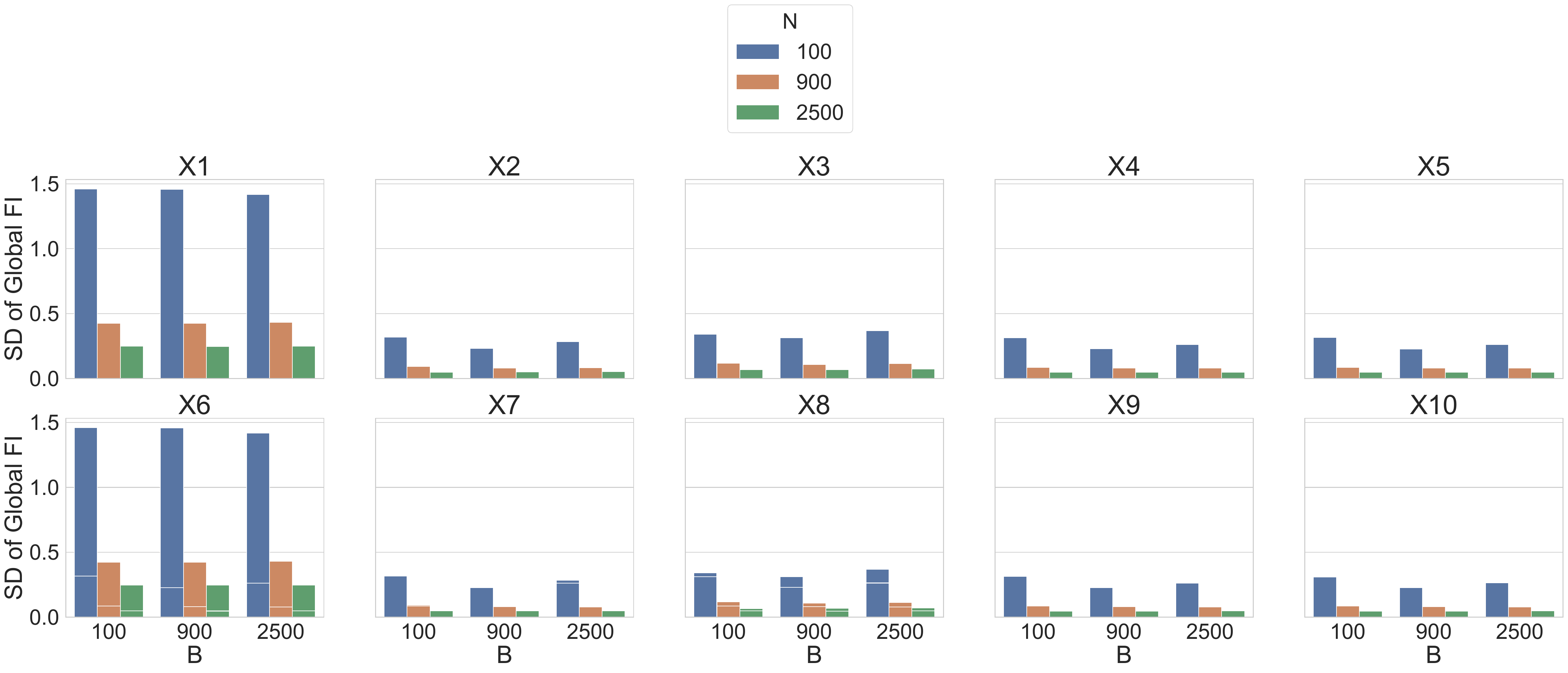}
     \caption{DGP-B}
     \end{subfigure}%
     \caption[]%
     {SD of global FI values of $p=10$ features for two functions; with respect to the number of permutations ($B$) and the number of observations ($N$).}
     \label{fig:pfi_var_sources}
\end{figure}

For both functions we compare the SD of the global FI values for different values of $B$ and $N$. In Figure \ref{fig:pfi_var_sources}, we can see that different features have different SDs, but in all conditions the SD is almost fixed with respect to $B$ and decreases with $N$. This indicates that the number of observations introduces more variability to the global FI values than the number of permutations.

\newpage
\section{Definitions and Proofs}\label{app:def_proof}

\subsection{Proof of Theorem \ref{thm}}
\label{app:proof}

In this section, we present the detailed proof of Theorem \ref{thm}.

Recall that $D$ is the set of partial rankings and that we assume that the probability of any error in $D$ is less than $\alpha$.
To prove the theorem, we first show that any error in coverage, i.e., a CI that does not cover the true rank, must be caused by at least one partial ranking error in $D$:

Suppose that there is a coverage error. Without loss of generality, assume that the coverage error occurs for feature 1: 
$$ [l_1, u_1] \nsubseteq [L_1,U_1].$$
The coverage error can be in one or both bounds:
\begin{enumerate}[label=(\arabic*)]
    \item $l_1 < L_1$,
    \item $u_1 > U_1$.
\end{enumerate}

If (1), then $L_1 > 1$, and there are $L_1-1>0$ pairs of the type $(1,k) \in D$, meaning that there are $L_1-1$ features with a significantly lower observed global FI than the observed global FI of feature 1.
However, according to Definition \ref{def:setrank}, $l_1-1 = \# \{k: \trueimp_1 > \trueimp_k\}$, meaning that there are only $l_1-1$ features with true global FI lower than the true global FI of feature 1.
Combining these two statements together, there must be at least one feature $k = 2,...,p$ for which 
$(1,k) \in D$ but $\trueimp_1 \ngtr \trueimp_k$, meaning that there is a partial ranking error in $D$.

If (2), then the set $\{k \in 2,...,p : (k,1) \in D\}$ is higher than the set $\{k: \trueimp_k>\trueimp_1\}$. Again, this would mean that for at least one value of $k$ there is a partial ranking error.

The event of at least one coverage error is contained in the event of obtaining a partial ranking error. Given that, the coverage error probability is bounded by $FWER = \alpha$.

\subsection{FWER Adjustment Procedures}\label{app:fwer}

Here, we provide details on the two sequential procedures that we use in our implementation. After adjustments, the p-values are compared to a chosen $\alpha$ level. Note that all p-values are inflated compared to their original level, making it less likely that the null hypothesis will be rejected. Furthermore, the p-values keep their relative order after adjustment. In the procedures below, this is governed by the $\max$ function, which assures that the order is maintained. The resulting process is sequential in that for a given level $\alpha$, after the first non-rejected value, all others would not be rejected. Let $p_1,...,p_K$ be a set of K p-values obtained by testing a family on null hypotheses $H_1^0, \ldots, H_0^K$; below we demonstrate how the two FWER adjustment procedures are used to calculate $p_1^{adg},...,p_K^{adg}$, a set of adjusted p-values.

\subsubsection{Holm's Procedure} 

We implement Holm's procedure \citep{holm1979simple} on one-sided hypothesis tests. The paired t-test is calibrated with this procedure for normally distributed base FI values.

Let $p_{(1)}\leq,...,\leq p_{(K)}\leq 1$ be the sorted set of p-values. Then:
\begin{align*}
    &p^{adj}_{(1)} = K\cdot p_{(1)}, \\
    &p^{adj}_{(2)} = \max\{p^{adj}_{(1)}, (K-1)p_{{(2)}}\}, \\
    &\ldots, \\
    &p^{adj}_{(k)} = \max\{p^{adj}_{(1)},...,p^{adj}_{(k-1)}, (K - (k-1))p_{(k)}\}, \\
    &\ldots, \\
    &p^{adj}_{(K)} = \max\{p^{adj}_{(1)},...,p^{adj}_{(K-1)}, p_{(K)}\}.
\end{align*}

\subsubsection{Min-P Procedure}

Holm's procedure is highly conservative, since it is valid regardless of the structure of dependence between the p-values. To improve it, \cite{westfall1993resampling} suggested the Min-P procedure. The idea is to use bootstrapping to model the structure of dependencies between p-values, obtain lower adjusted p-values, and reject more hypotheses. The details of the Min-P procedure described here are taken from \cite{efron2012large}. 

Here, we also start with the sorted set of p-values $p_{(1)}\leq,...,\leq p_{(K)}\leq 1$. Let $i_1, \ldots i_K$ indicate the corresponding original indices, $p_{(k)} = p_{i_k}$, and define $I_k=\{ i_k, i_{k+1}, \ldots, i_K \}$ and $\pi_k=\mathcal{P}_0 \{ \min_{j \in I_k} (P_j) \leq p_{(k)}\}$. Here, $(P_1, \ldots, P_K)$ indicates a hypothetical realization of the unordered p-values $p_1,...,p_K$ obtained under the complete null hypothesis, meaning all $H_k^0$s are true. 
The adjusted p-values are then defined by:
\begin{align*}
    &p^{adj}_j = \max_{k \leq j} \pi_k.
\end{align*}

\newpage
\section{Confident Feature Ranking: Step-By-Step}\label{app:full_example}

This section demonstrates how our ranking method constructs simultaneous CIs for the true ranks. We use the bike sharing dataset \citep{fanaee2014event}, the TreeSHAP FI method, and our ranking method with Holm's procedure for $n=50$ base FI values. This demonstration is a detailed description of the example presented in Section \ref{sec:real} (Figure \ref{fig:bike_base_size}).

\paragraph{Base and Global FI Values} We construct a TreeSHAP explainer \citep{lundberg2019Tree} based on the trained XGB model. The base FI values are the absolute values of the local SHAP values the explainer produces for an explanation set of size $n=50$. The global FI values are the average of the base FI values. The values and the order of the global base FI values for this explanation set are presented in Table \ref{tab:full_example}.

\begin{figure}[ht]
     \centering
     \begin{subfigure}{0.3\columnwidth}
     \centering
     \includegraphics[width=\columnwidth]{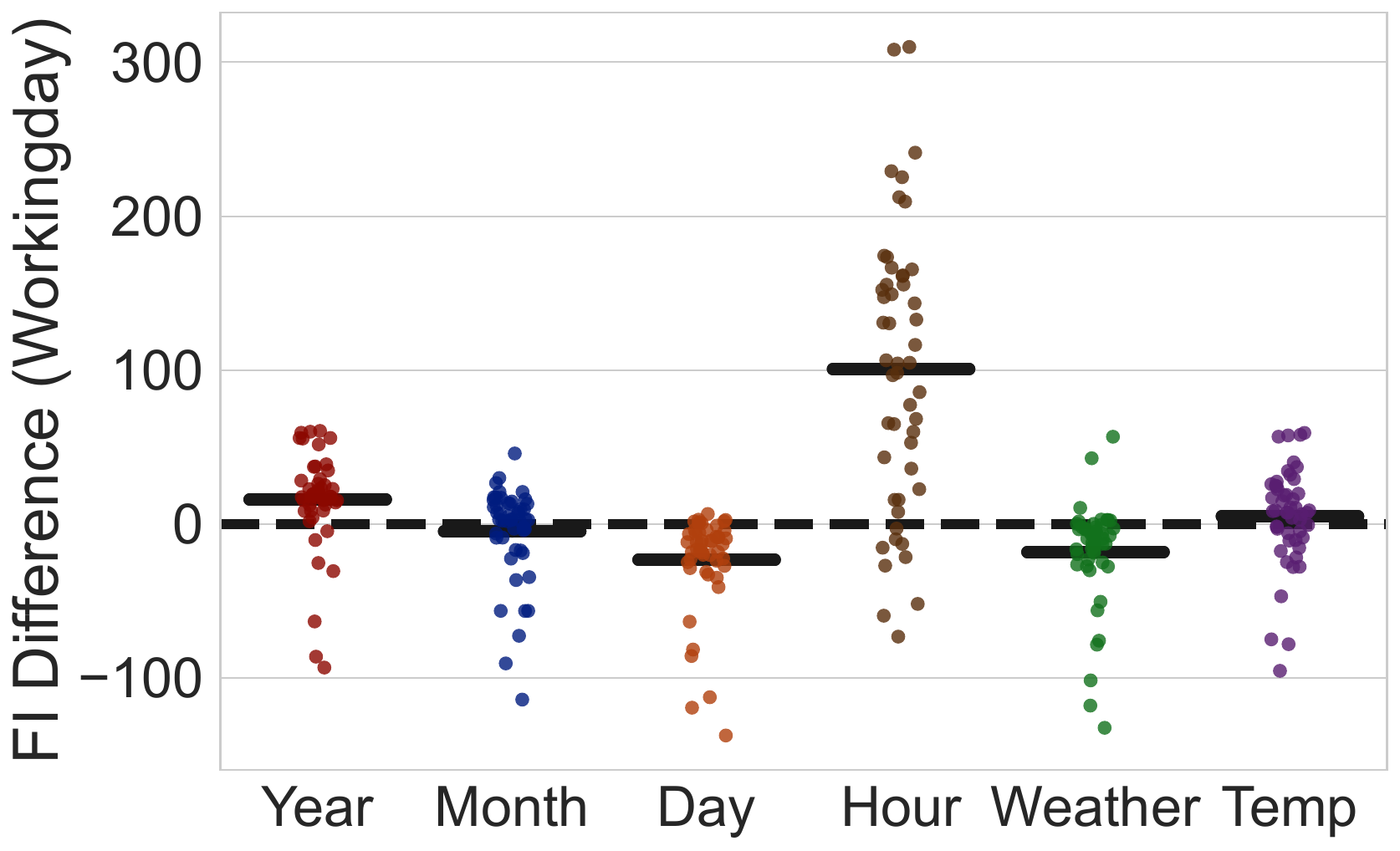}
     \caption{Base-FI differences}
     \label{fig:pair_diff}
     \end{subfigure}%
     \begin{subfigure}{0.3\columnwidth}
     \centering
     \includegraphics[width=\columnwidth]{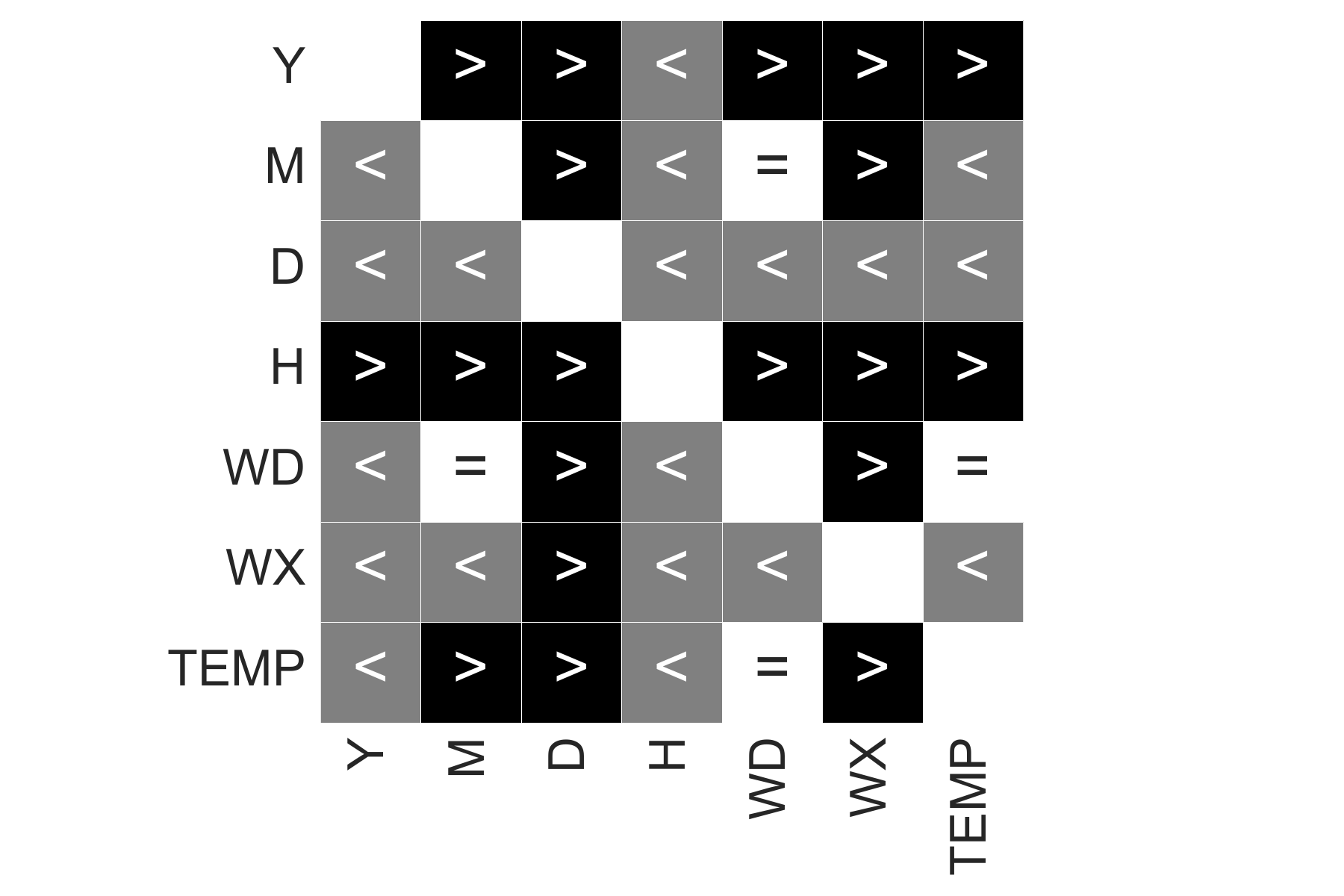}
     \caption{Partial rankings}
     \label{fig:signs}
     \end{subfigure}%
     \begin{subfigure}{0.3\columnwidth}
     \centering
     \includegraphics[width=\columnwidth]{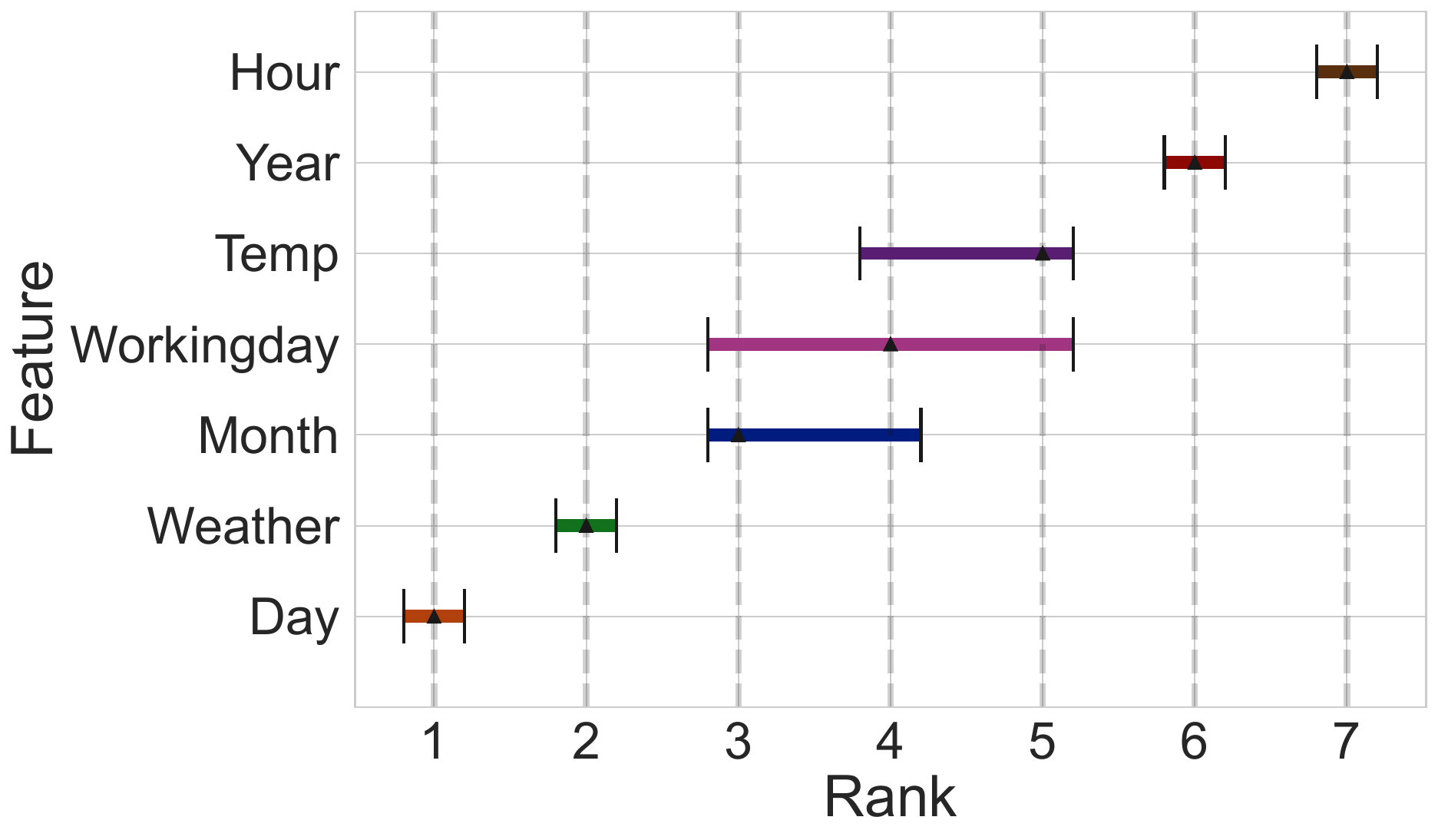}
     \caption{CIs for the true ranks}
     \label{fig:full_example_rank}
     \end{subfigure}%
     \caption[]%
     {Visualization of Algorithm \ref{alg}: first, test all one-sided pairs of hypotheses for the difference between base FI values (a), then adjust the p-values and obtain the partial rankings from the rejected hypotheses (b), and finally construct the CIs for the true ranks (c).}
     \label{fig:full_example}
\end{figure}

\paragraph{Pairwise Differences} The paired-sample t-test is based on the differences between the base FI values of two features $\basematrix_j - \basematrix_k$. The one-sided hypothesis $H_{jk}^0: \trueimp_j \geq \trueimp_k$ is rejected if the difference between the observed global FI values is significantly different from zero. In Figure \ref{fig:pair_diff}, we present the differences between \emph{Workingday} and all other features. The average of the differences between \emph{Workingday} and \emph{Month} and \emph{Temp} is near zero.

\paragraph{Partial Rankings} We set the significance level to $\alpha=0.1$, and for each pair of features, we run two paired one-sided t-tests; then, we adjust the p-values to multiple comparisons using Holm's procedure. In Figure \ref{fig:signs}, gray and black indicate that the observed global FI value of the feature in row $j$ is respectively less and greater than the observed global FI value of the feature in column $k$. White indicates that the difference is zero (neither $H_{jk}^0$  nor $H_{kj}^0$ were rejected). The set of partial rankings $\paranking$ is then obtained. For example, we can conclude that $(Month, Year) \in \paranking$, $(Day, Month) \in \paranking$, and $(Month, Workingday) \not\in \paranking$.

\paragraph{Constructing Simultaneous CIs for the True Ranks}
For each feature, we initialize the lower bound of the CI to one and the upper bound to $p$.
If there are no differences between the features, the CIs for all features are $[1, p]$. Otherwise, there are differences. Without loss of generality, consider the \emph{Workingday} feature. By looking at the row for \emph{Workingday} in Figure \ref{fig:signs}, we can see that the observed global FI value of \emph{Workingday} is significantly higher than the observed global FI values of \emph{Day} and \emph{Weather}, and it is significantly lower than the observed global FI values of \emph{Year} and \emph{Hour}. There is no significant difference between \emph{Workingday} and \emph{Month} and \emph{Temp}. Therefore, we increase the lower bound by two, decrease the upper bound by two, and obtain the confidence set $[3, 5]$ for the true rank of \emph{Workingday}.

We repeat the same process for all features and obtain $90\%$ simultaneous CIs for the true ranks. See lower and upper bounds in Table \ref{tab:full_example} and a visualization of the CIs in Figure \ref{fig:full_example_rank}.

\begin{table}[ht]
\centering
\caption{Ranks and Simultaneous CIs}
\label{tab:full_example}
\begin{tabular}{|l|c|c|c|}
\hline
\textbf{Feature} & \multicolumn{1}{l|}{\textbf{Observed Global FI}} & \multicolumn{1}{l|}{\textbf{Observed Rank}} & \multicolumn{1}{l|}{\textbf{CI}} \\ \hline
Hour             & 129.042                                          & 7                                           & [7, 7]                           \\ \hline
Year             & 44.805                                           & 6                                           & [6, 6]                           \\ \hline
Temp             & 33.777                                           & 5                                           & [4, 5]                           \\ \hline
Workingday       & 28.95                                            & 4                                           & [3, 5]                           \\ \hline
Month            & 23.987                                           & 3                                           & [3, 4]                           \\ \hline
Weather          & 10.865                                           & 2                                           & [2, 2]                           \\ \hline
Day              & 5.673                                            & 1                                           & [1, 1]                           \\ \hline
\end{tabular}
\end{table}

\newpage
\section{Experiment Details and Additional Results}\label{app:exp}

\subsection{Ranking Method Comparison}\label{app:mock}

\subsubsection{Baseline Ranking Method}\label{app:naive}
We implement a naive ranking method to construct CIs for the features' true ranks based on bootstrap samples. For each sample, we rank the global FI values. We report lower and upper bounds ($L_j, U_j$) by taking the $\alpha/2$ and $1 - \alpha/2$ quantiles of the ranks of the $j$'th feature over the bootstrap distribution.

\subsubsection{Additional Results}

\begin{figure}[ht]
     \centering
     \begin{subfigure}{0.49\columnwidth}
     \centering
     \includegraphics[width=\columnwidth]{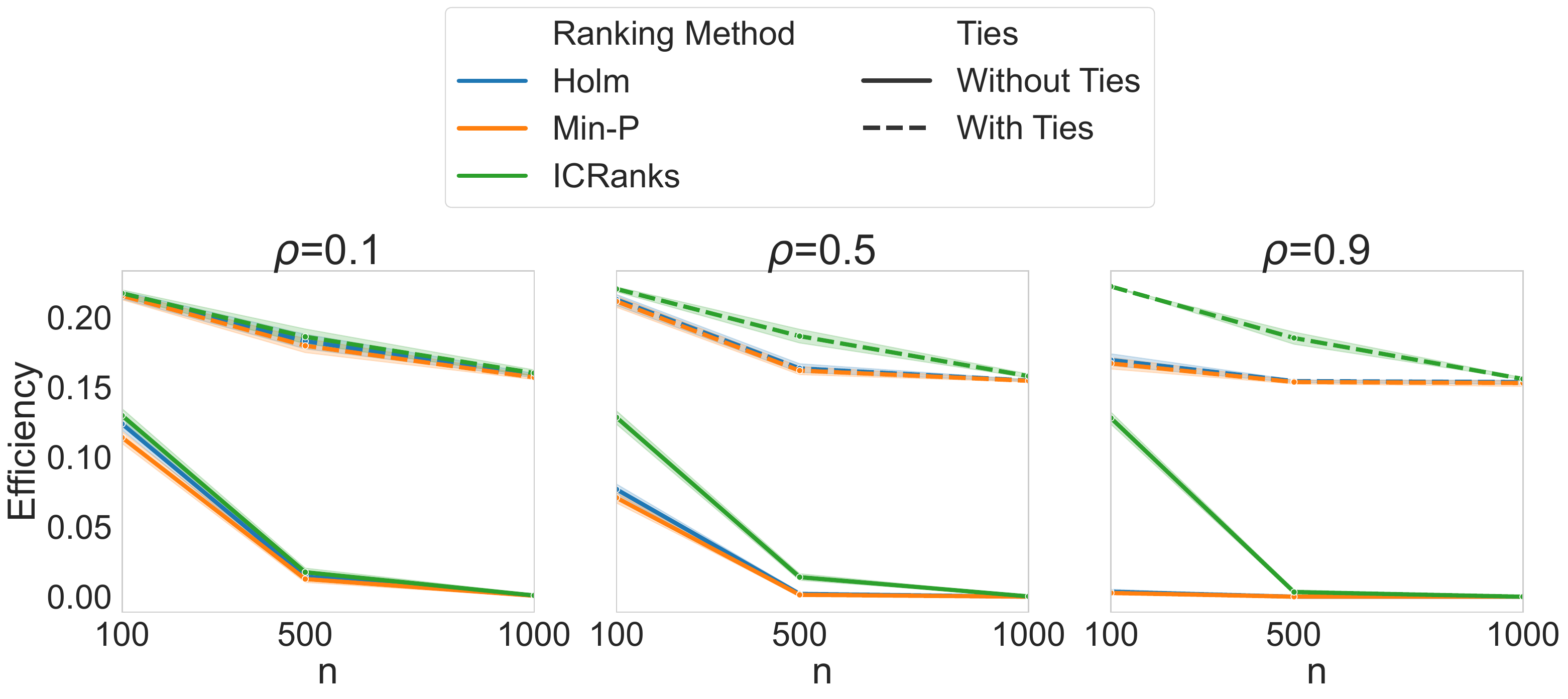}
     \caption{$p=10$: $\sigma$-factor=0.2}
     \end{subfigure}%
     \begin{subfigure}{0.49\columnwidth}
     \centering
     \includegraphics[width=\columnwidth]{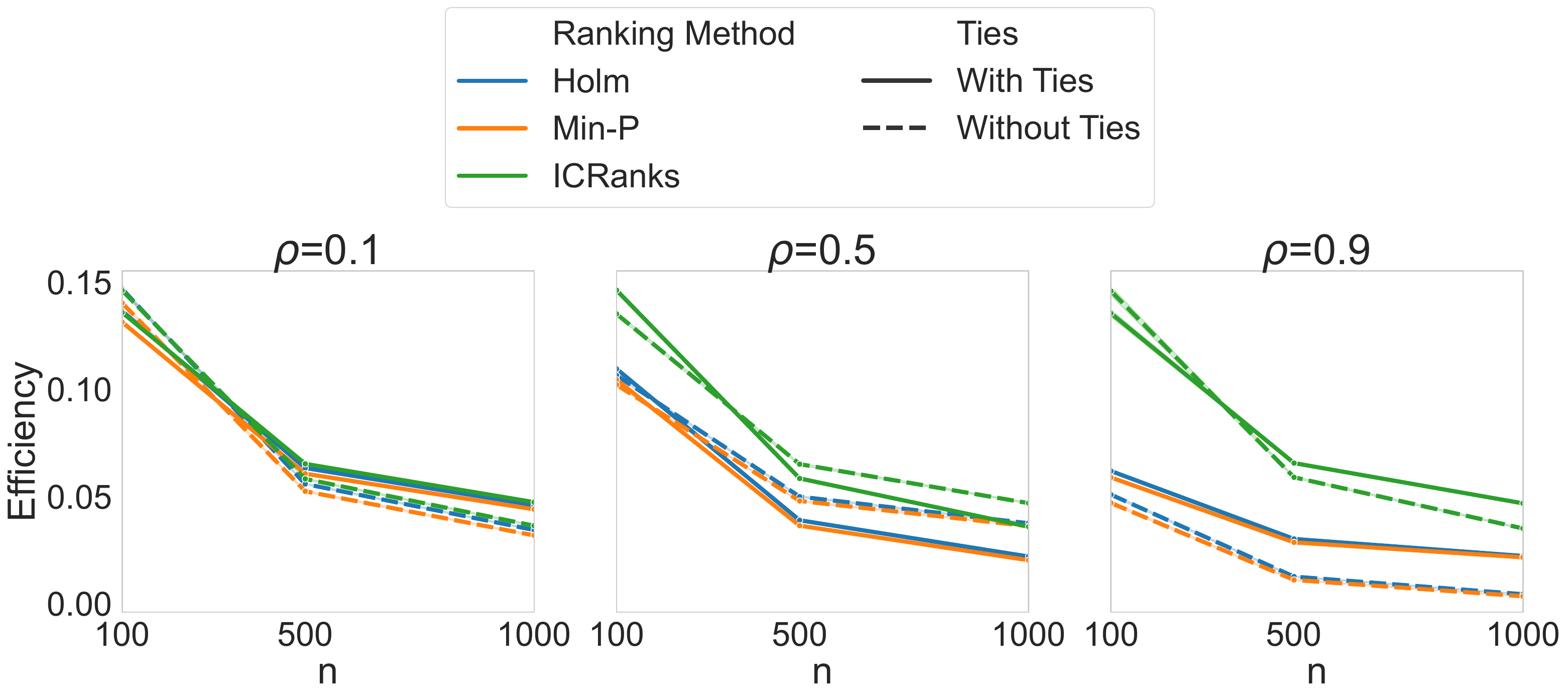}
     \caption{$p=50$: $\sigma$-factor=0.2}
     \end{subfigure}%
     \\
     \begin{subfigure}{0.49\columnwidth}
     \centering
     \includegraphics[width=\columnwidth]{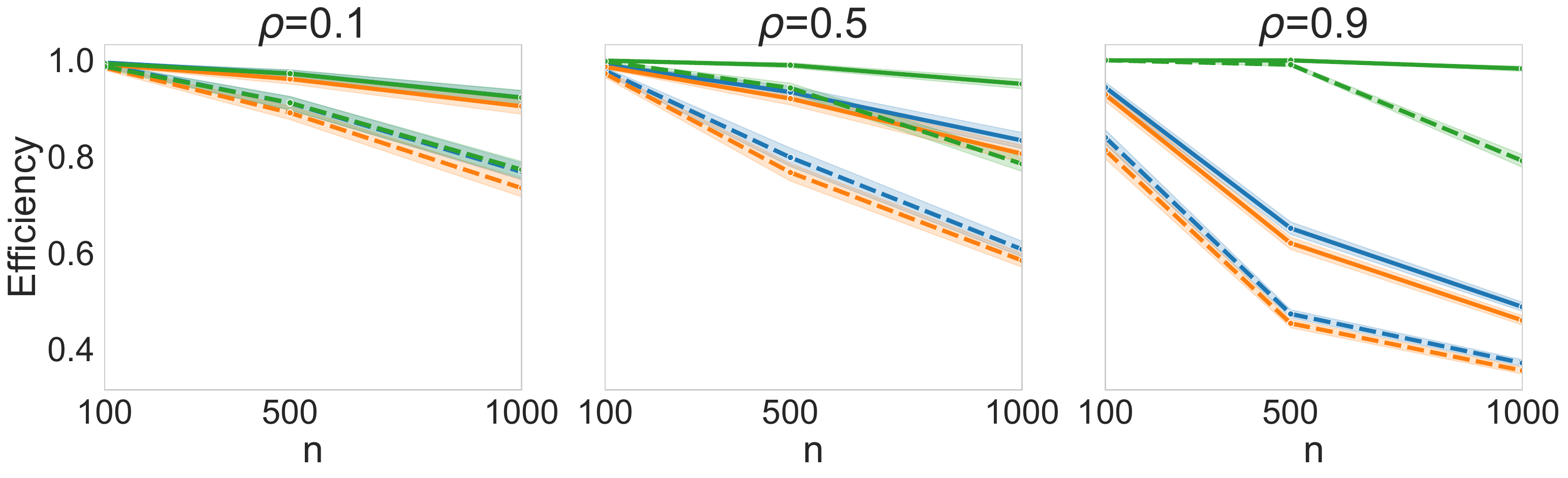}
     \caption{$p=10$: $\sigma$-factor=5}
     \end{subfigure}%
     \begin{subfigure}{0.49\columnwidth}
     \centering
     \includegraphics[width=\columnwidth]{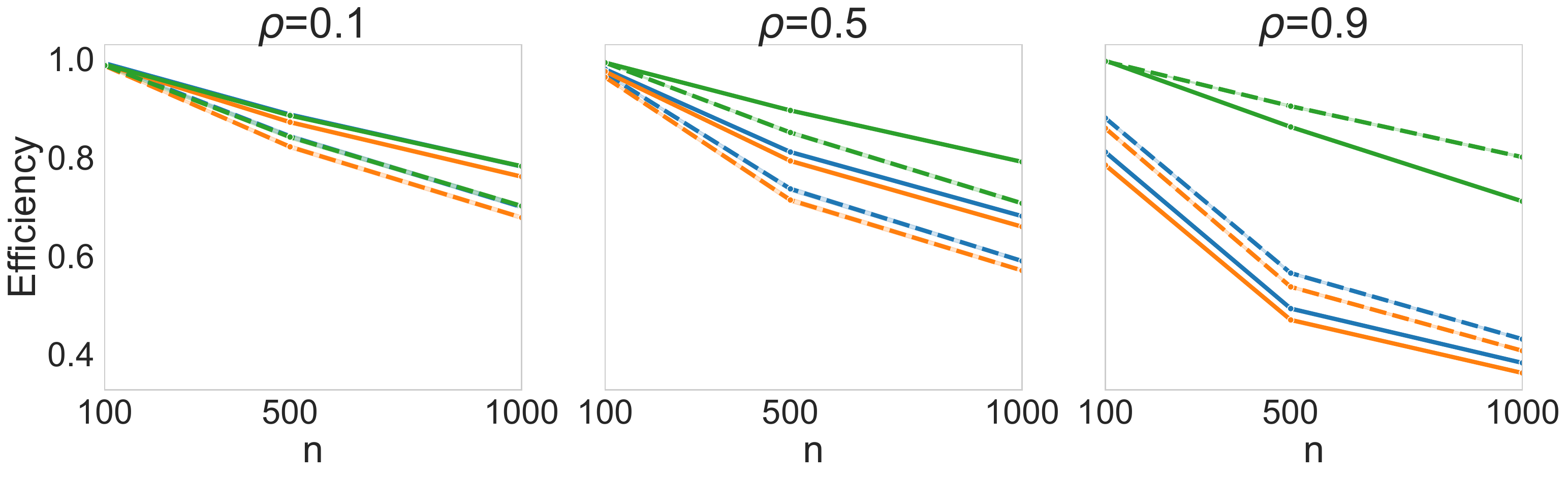}
     \caption{$p=50$: $\sigma$-factor=5}
     \end{subfigure}%
     \caption[]%
     {Ranking efficiency for $p=10$ and $p=50$.}
     \label{fig:mock_exp_corr_10_50}
\end{figure}

\paragraph{Equal Correlations} In Figure \ref{fig:mock_exp_corr_10_50}, we present the ranking efficiency of three ranking methods for $p=10$ and $p=50$ as a function of $n$, with multiple levels of correlations. We use the same configuration as in Figure \ref{fig:mock_exp_corr}: low and high $\sigma$-factors, $\mu$-exponent=0.25, with and without ties. We can observe the same efficiency trends seen for $p=30$: 
the efficiency increases as $\rho$ increases, the gap between the methods increases as the $\sigma$-factor increases, the efficiency improves as the $n$ increases, and our ranking method is more efficient than ICRanks.

\begin{figure}[ht]
    \centering\includegraphics[width=0.5\columnwidth]{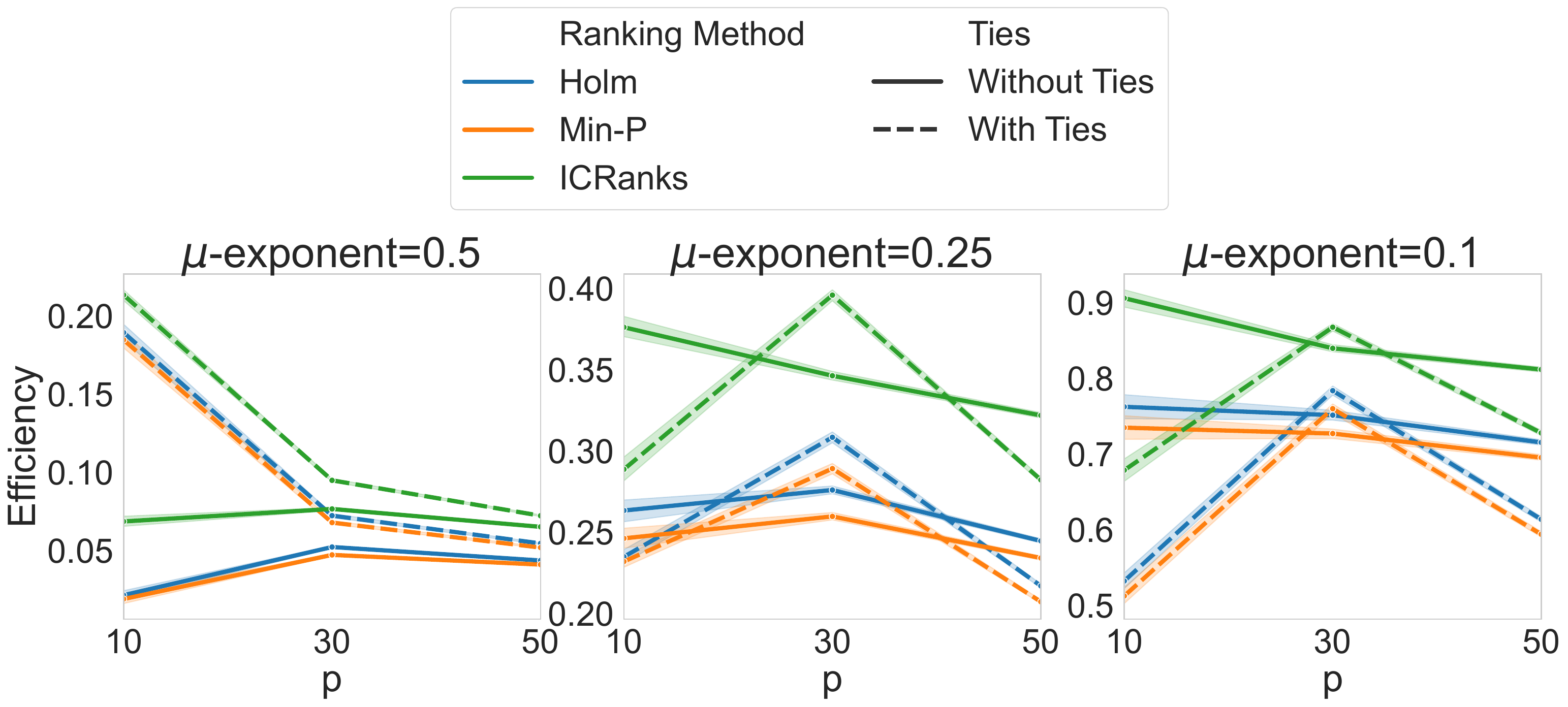}
    \caption[]%
     {Ranking efficiency as a function of $p$ for multiple values of $\mu$-exponent and three ranking methods.}
    \label{fig:mock_compare_p}
\end{figure}

\paragraph{Number of Features} We compare the ranking efficiency for different numbers of features and multiple values of $\mu$-exponent, with $n=500$, $\sigma$-factor=1, equal correlations, and $\rho=0.5$ (see Figure \ref{fig:mock_compare_p}). The efficiency degrades as the means become more dense ($\mu$-exponent decreases), , and the number of features has almost no effect on the efficiency.

\begin{figure}[ht]
     \centering
     \begin{subfigure}{0.35\columnwidth}
     \centering
     \includegraphics[width=\columnwidth]{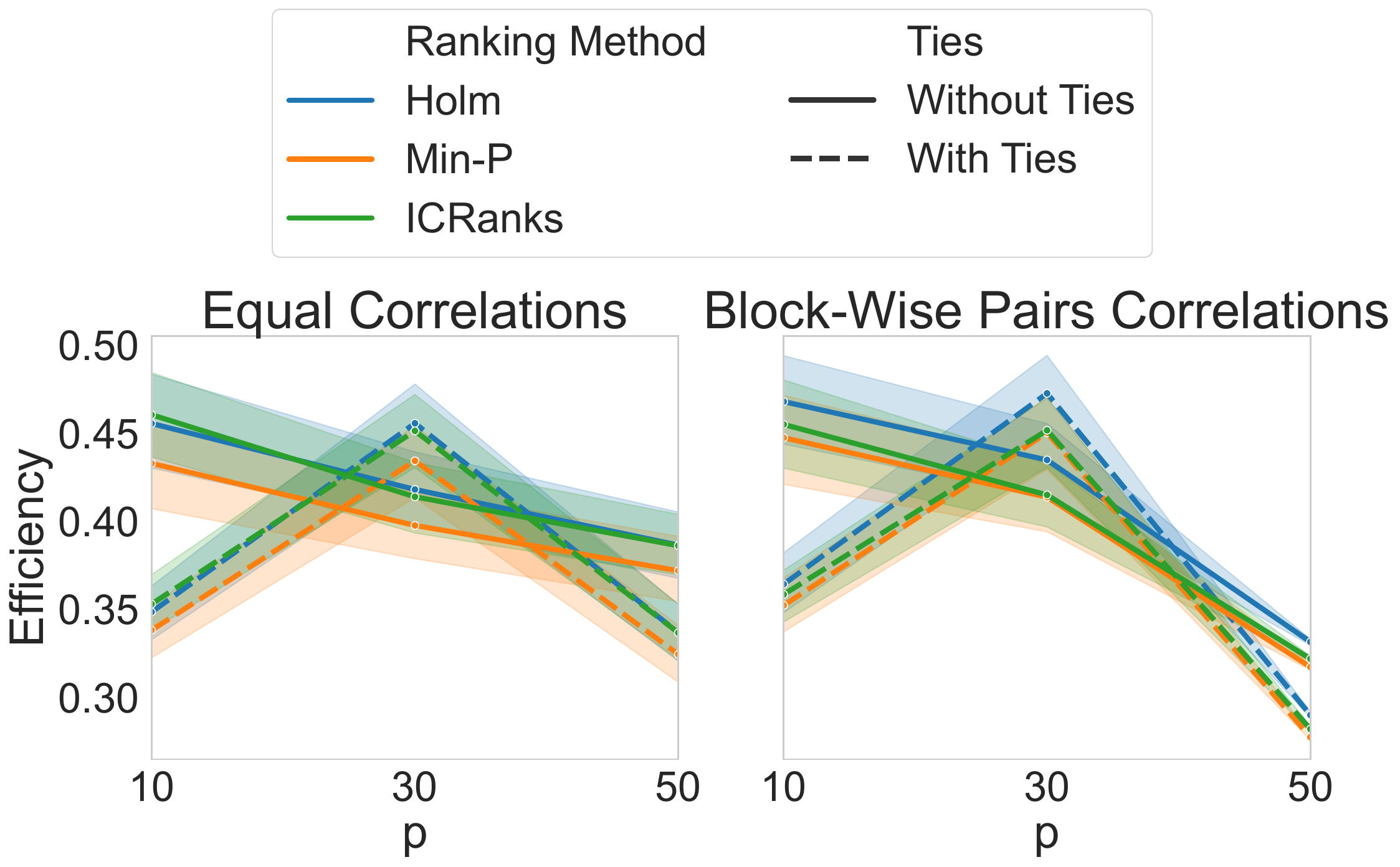}
     \caption{$\rho$=0.1}
     \end{subfigure}%
     \\
     \begin{subfigure}{0.35\columnwidth}
     \centering
     \includegraphics[width=\columnwidth]{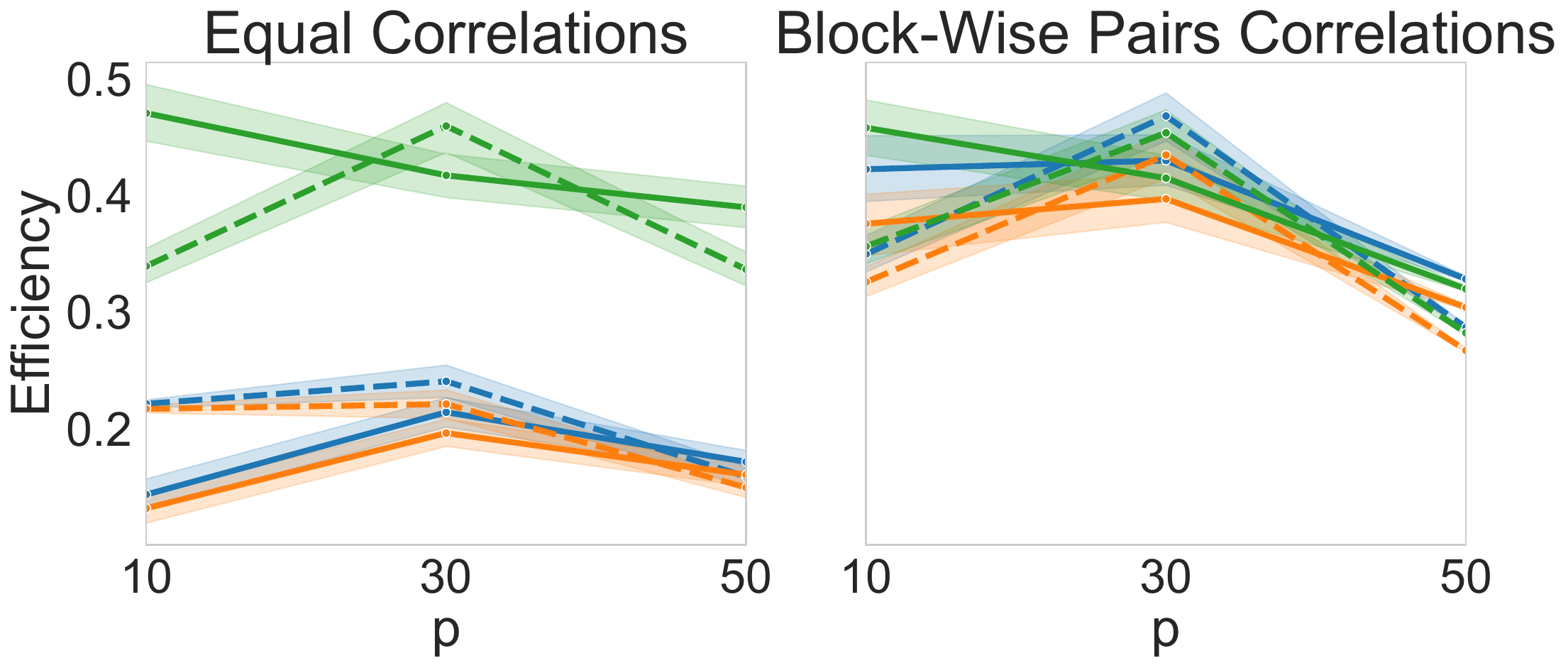}
     \caption{$\rho$=0.9}
     \end{subfigure}%
     \caption[]%
     {Ranking efficiency with low (a) and high (b) values of $\rho$, as a function of $p$ for two correlation structures and three ranking methods.}
     \label{fig:mock_compare_corr}
\end{figure}

\paragraph{Correlation Structure} We compare the equal correlation structure with the block-wise pairs structure of the correlation matrix. In Figure \ref{fig:mock_compare_corr}, we present the results for different numbers of features, $n=500$, $\sigma$-factor=1, $\mu$-exponent-0.25, and low (0.1) and high (0.9) values of $\rho$. The differences between the structures of the correlation matrix are more substantial for $\rho=0.9$.

\subsection{SHAP Ranking Measures}\label{app:shap_simulated}

Here, we sample the training and explanation sets using the DGP described in Section \ref{sec:shap_simulated}. For each configuration of $(X, Y)$, we train an XGB (default hyperparameters) or RF (number of estimators=1,000) regression model for this experiment. We follow the XGB tutorial\footnote{\href{https://xgboost.readthedocs.io/en/stable/tutorials/rf.html}{XGB tutorial}} to train both the XGB and RF models (see the train and test $R^2$ of the model in Table \ref{tab:shap_r2}).

\begin{table}[ht]
\centering
\caption{Prediction Models' Performance}
\label{tab:shap_r2}
\begin{tabular}{|l|c|c|c|c|}
\hline
\textbf{Model}       & \multicolumn{1}{l|}{\textbf{DGP}} & \multicolumn{1}{l|}{\textbf{p}} & \multicolumn{1}{l|}{\textbf{Train $R^2$}} & \multicolumn{1}{l|}{\textbf{Test $R^2$}} \\ \hline
\multirow{6}{*}{RF}  & \multirow{3}{*}{DGP-A}           & 10                              & 0.786                                     & 0.783                                    \\ \cline{3-5} 
                     &                                        & 30                              & 0.523                                     & 0.515                                    \\ \cline{3-5} 
                     &                                        & 50                              & 0.346                                     & 0.336                                    \\ \cline{2-5} 
                     & \multirow{3}{*}{DGP-B}           & 10                              & 0.863                                     & 0.862                                    \\ \cline{3-5} 
                     &                                        & 30                              & 0.836                                     & 0.836                                    \\ \cline{3-5} 
                     &                                        & 50                              & 0.745                                     & 0.744                                    \\ \hline
\multirow{6}{*}{XGB} & \multirow{3}{*}{DGP-A}           & 10                              & 0.956                                     & 0.952                                    \\ \cline{3-5} 
                     &                                        & 30                              & 0.963                                     & 0.958                                    \\ \cline{3-5} 
                     &                                        & 50                              & 0.954                                     & 0.945                                    \\ \cline{2-5} 
                     & \multirow{3}{*}{DGP-B}           & 10                              & 0.875                                     & 0.868                                    \\ \cline{3-5} 
                     &                                        & 30                              & 0.951                                     & 0.946                                    \\ \cline{3-5} 
                     &                                        & 50                              & 0.964                                     & 0.959                                    \\ \hline
\end{tabular}
\end{table}

\subsubsection{Additional Results}

\begin{figure}[ht]
     \centering
     \begin{subfigure}{0.5\columnwidth}
     \centering
     \includegraphics[width=\columnwidth]{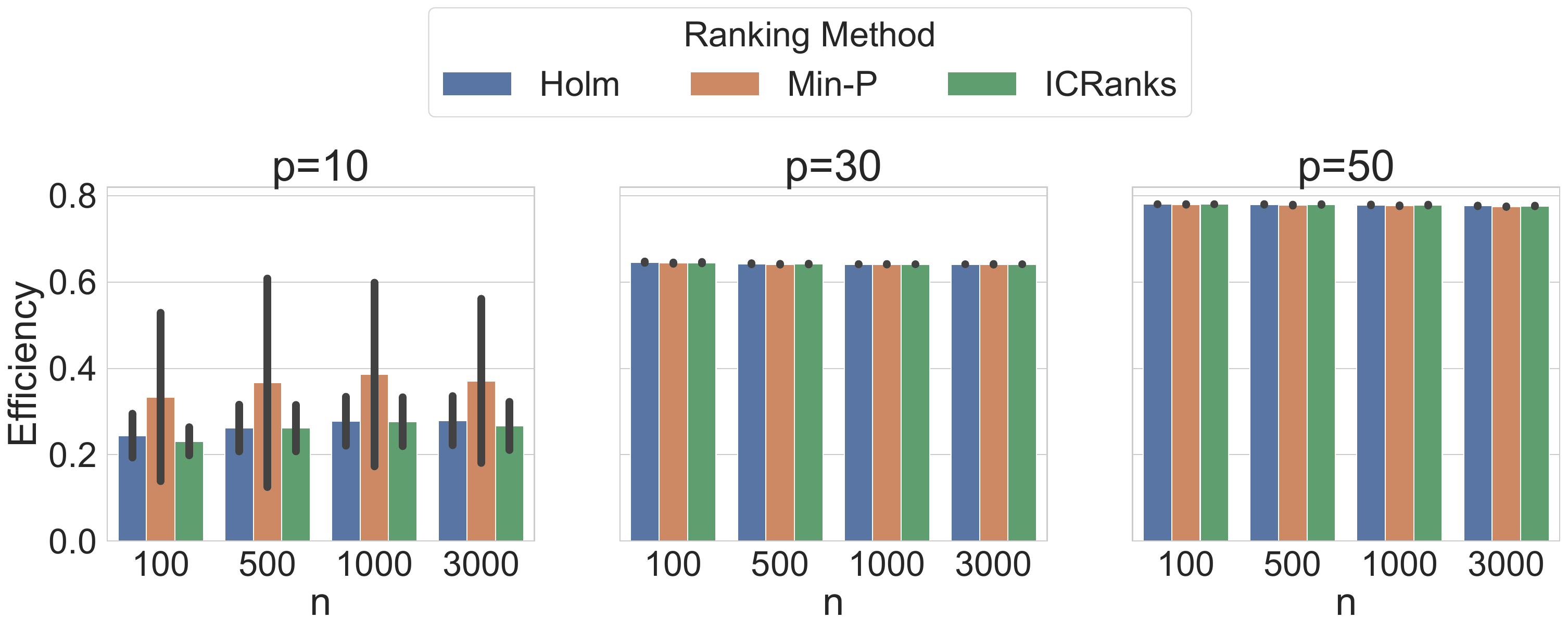}
     \caption{RF with DGP-B}
     \end{subfigure}%
     \\
     \begin{subfigure}{0.5\columnwidth}
     \centering
     \includegraphics[width=\columnwidth]{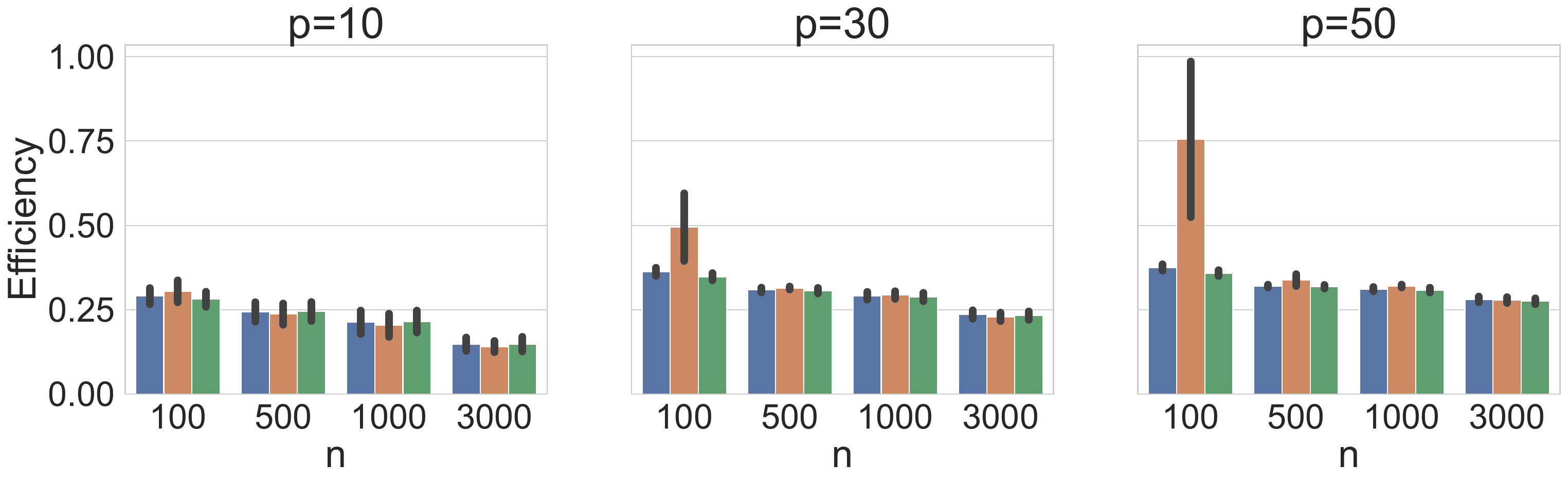}
     \caption{XGB with DGP-A}
     \end{subfigure}%
     \caption[]%
     {Ranking efficiency as a function of $n$ for different numbers of features ($p$) and ranking methods.}
     \label{fig:shap_additional}
\end{figure}

In the paper, we present an example of the efficiency of RF with DGP-A and XGB with DGP-B. Here, we present the complementary efficiency results for all configurations (see Figure \ref{fig:shap_additional}). The simultaneous coverage for all configurations is almost one ($0.997 \pm 0.009$). In addition, we compare the efficiency of our CIs (using our method with the Min-P procedure) with $n=1000$ base FI values, to the efficiency of the true FI ranks as an upper bound on the efficiency of the observed values (see Table \ref{tab:true_efficiency}); as can be seen, the efficiency of our ranking method with $n=1000$ is not ideal, even in the case of perfect true ranking.

\begin{table}[ht]
\centering
\caption{Ranking Efficiency}
\label{tab:true_efficiency}
\begin{tabular}{|l|c|c|c|c|}
\hline
\textbf{Model}       & \multicolumn{1}{l|}{\textbf{DGP}} & \multicolumn{1}{l|}{\textbf{p}} & \multicolumn{1}{l|}{\textbf{True Efficiency}} & \multicolumn{1}{l|}{\textbf{Mean Efficiency}} \\ \hline
\multirow{6}{*}{RF}  & \multirow{3}{*}{DGP-A}           & 10                              & 0.222                                         & 0.238                                         \\ \cline{3-5} 
                     &                                        & 30                              & 0.393                                         & 0.412                                         \\ \cline{3-5} 
                     &                                        & 50                              & 0.486                                         & 0.495                                         \\ \cline{2-5} 
                     & \multirow{3}{*}{DGP-B}           & 10                              & 0.222                                             & 0.386                                         \\ \cline{3-5} 
                     &                                        & 30                              & 0.634                                         & 0.641                                         \\ \cline{3-5} 
                     &                                        & 50                              & 0.772                                         & 0.777                                         \\ \hline
\multirow{6}{*}{XGB} & \multirow{3}{*}{DGP-A}           & 10                              & 0                                             & 0.204                                         \\ \cline{3-5} 
                     &                                        & 30                              & 0                                             & 0.293                                         \\ \cline{3-5} 
                     &                                        & 50                              & 0                                             & 0.32                                          \\ \cline{2-5} 
                     & \multirow{3}{*}{DGP-B}           & 10                              & 0                                             & 0.299                                         \\ \cline{3-5} 
                     &                                        & 30                              & 0                                             & 0.247                                         \\ \cline{3-5} 
                     &                                        & 50                              & 0                                             & 0.287                                         \\ \hline
\end{tabular}
\end{table}

\subsubsection{Ranking Runtime Analysis}\label{app:shap_simulated_runtime}

\begin{figure}[ht]
     \centering
     \begin{subfigure}{0.5\columnwidth}
     \centering
     \includegraphics[width=\columnwidth]{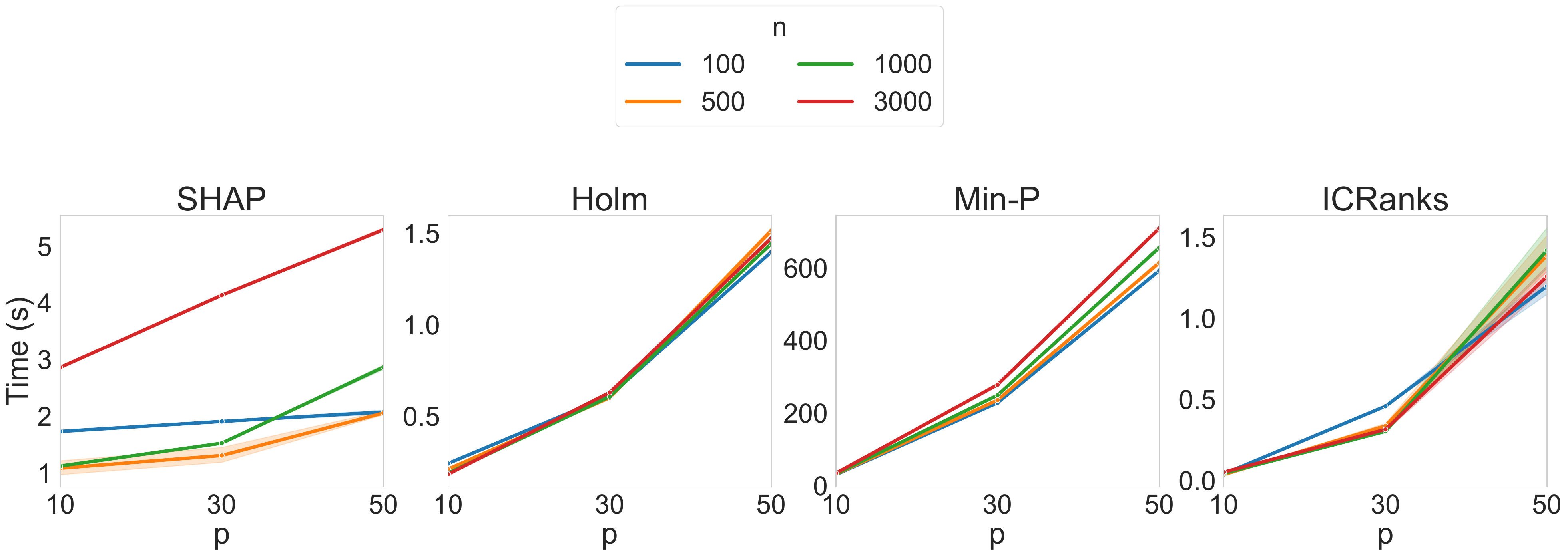}
     \caption{RF with DGP-A}
     \end{subfigure}%
     \\
     \begin{subfigure}{0.5\columnwidth}
     \centering
     \includegraphics[width=\columnwidth]{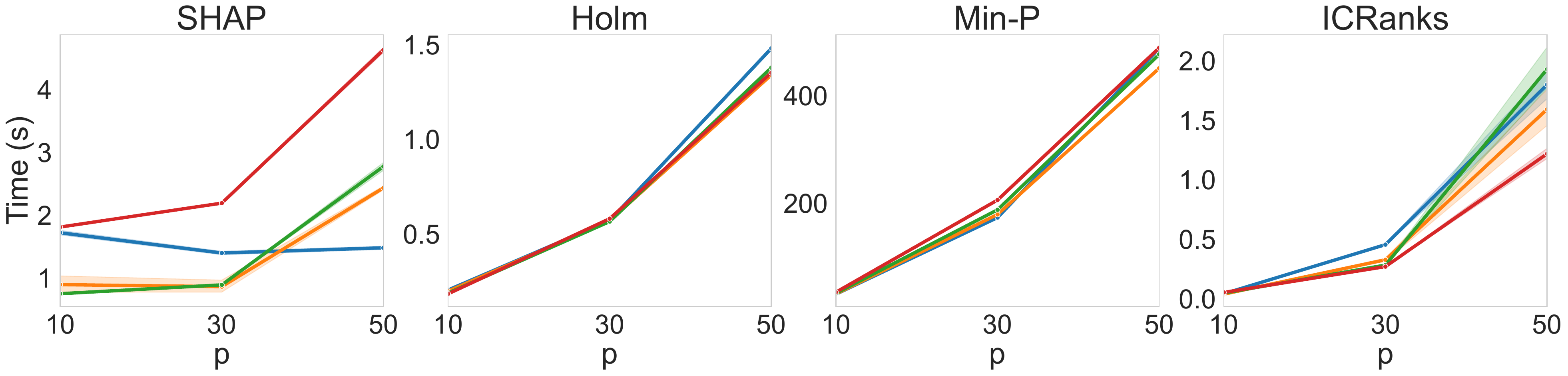}
     \caption{RF with DGP-B}
     \end{subfigure}%
     \\
     \begin{subfigure}{0.5\columnwidth}
     \centering
     \includegraphics[width=\columnwidth]{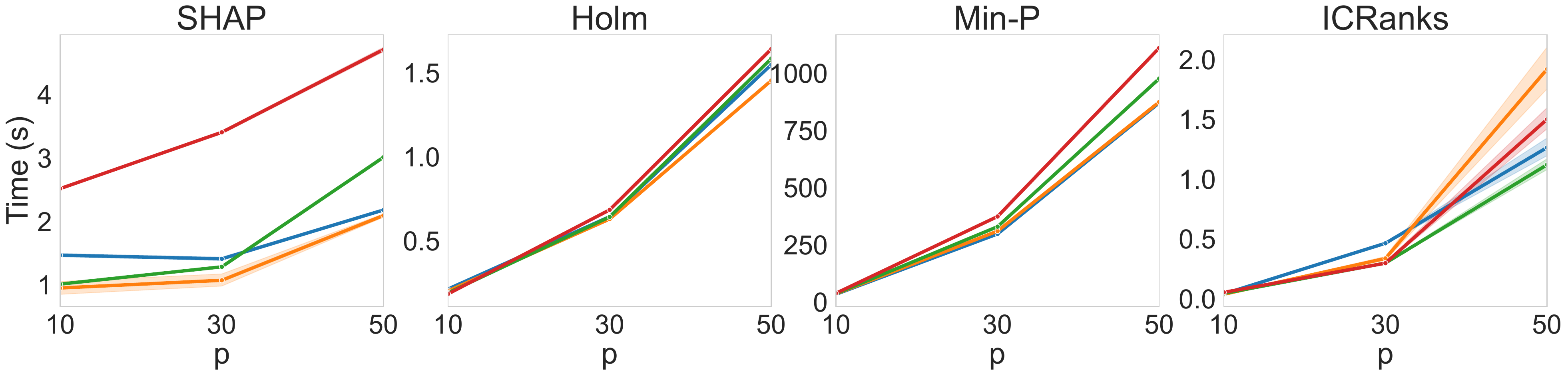}
     \caption{XGB with DGP-A}
     \end{subfigure}%
     \\
     \begin{subfigure}{0.5\columnwidth}
     \centering
     \includegraphics[width=\columnwidth]{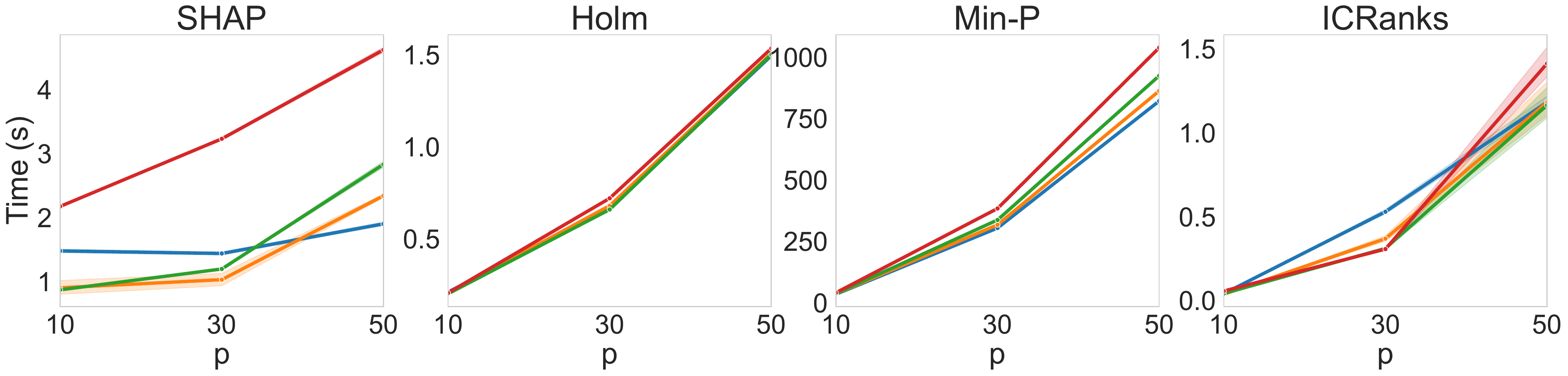}
     \caption{XGB with DGP-B}
     \end{subfigure}%
     \caption[]%
     {TreeSHAP and ranking times (in seconds) as a function of the number of features ($p$) for different sizes ($n$) of $\sampleset_{explain}$.}
     \label{fig:shap_times}
\end{figure}

We also analyzed TreeSHAP's runtime and our method's runtime. In Figure \ref{fig:shap_times}, we present the runtime of TreeSHAP and the different ranking methods. TreeSHAP's runtime clearly depends on the sample size $n$, since it is a local FI method. In contrast, the runtime of the ranking methods depends on the number of features ($p$). The runtime of our method with Holm's procedure is comparable to that of ICRanks but with lower variance. The runtime of our method with the Min-P procedure is much higher, because it is based on a bootstrap process and increases with the number of repetitions ($B$).

\subsubsection{Non-Normal Base FI Value Distribution}\label{app:shap_simulated_non_normal}

We use the paired-sample t-test to compare base FI values and adjust the p-values with the Min-P or Holm's procedure. Our primary assumption is that the paired test is calibrated for the possible distributions of base FI values (note that the paired-sample t-test is calibrated even when the base FI values are not normally distributed). However, we found that our method does not always maintain simultaneous coverage when the base FI values have an extremely long tail. In this example, we sample the data from:
\begin{equation*}
\begin{split}
    y &= x_1x_2 + x_3^2 - x_4x_7 + x_8x_{10} - x_6^2 \\
    & + x_{11}x_{12} + x_{13}^2 - x_{14}x_{17} + x_{18}x_{20} - x_{16}^2 \\
    & + x_{21}x_{22} + x_{23}^2 - x_{24}x_{27} + x_{28}x_{30} - x_{26}^2 + \epsilon; \\
    & X \sim N_{30}(\mathbf{0}, \Sigma); \, \epsilon \sim N(0, 1), \\
    & \text{where } \Sigma \text{ is an equal correlation matrix} \\
    & \text{with } \rho=0.3 \text{ and } \{\sigma_j^2\} \sim \chi^2.
\end{split}
\end{equation*}

We train an RF model and calculate the base FI values with TreeSHAP. Table \ref{tab:non_normal} summarizes the average coverage and simultaneous coverage. As can be seen, for all sizes of $n$ (the number of base FI values) our method does not maintain simultaneous coverage; the marginal coverage is almost $90\%$ for small sizes of $n$, and the simultaneous coverage of the Min-P procedure is better.

\begin{table}[ht]
\centering
\caption{Ranking Coverage}
\label{tab:non_normal}
\begin{tabular}{|l|c|c|c|}
\hline
\textbf{n}            & \multicolumn{1}{l|}{\textbf{Ranking Method}} & \multicolumn{1}{l|}{\textbf{Coverage}} & \multicolumn{1}{l|}{\textbf{Simultaneous Coverage}} \\ \hline
\multirow{2}{*}{100}  & Holm                                         & 0.839                                  & 0.08                                                \\ \cline{2-4} 
                      & Min-P                                        & 0.96                                   & 0.63                                                \\ \hline
\multirow{2}{*}{500}  & Holm                                         & 0.75                                   & 0.01                                                \\ \cline{2-4} 
                      & Min-P                                        & 0.888                                  & 0.56                                                \\ \hline
\multirow{2}{*}{1000} & Holm                                         & 0.7                                    & 0.01                                                \\ \cline{2-4} 
                      & Min-P                                        & 0.85                                   & 0.37                                                \\ \hline
\multirow{2}{*}{3000} & Holm                                         & 0.634                                  & 0.01                                                \\ \cline{2-4} 
                      & Min-P                                        & 0.782                                  & 0.14                                                \\ \hline
\end{tabular}
\end{table}

\begin{figure}[ht]
    \centering\includegraphics[width=0.4\columnwidth]{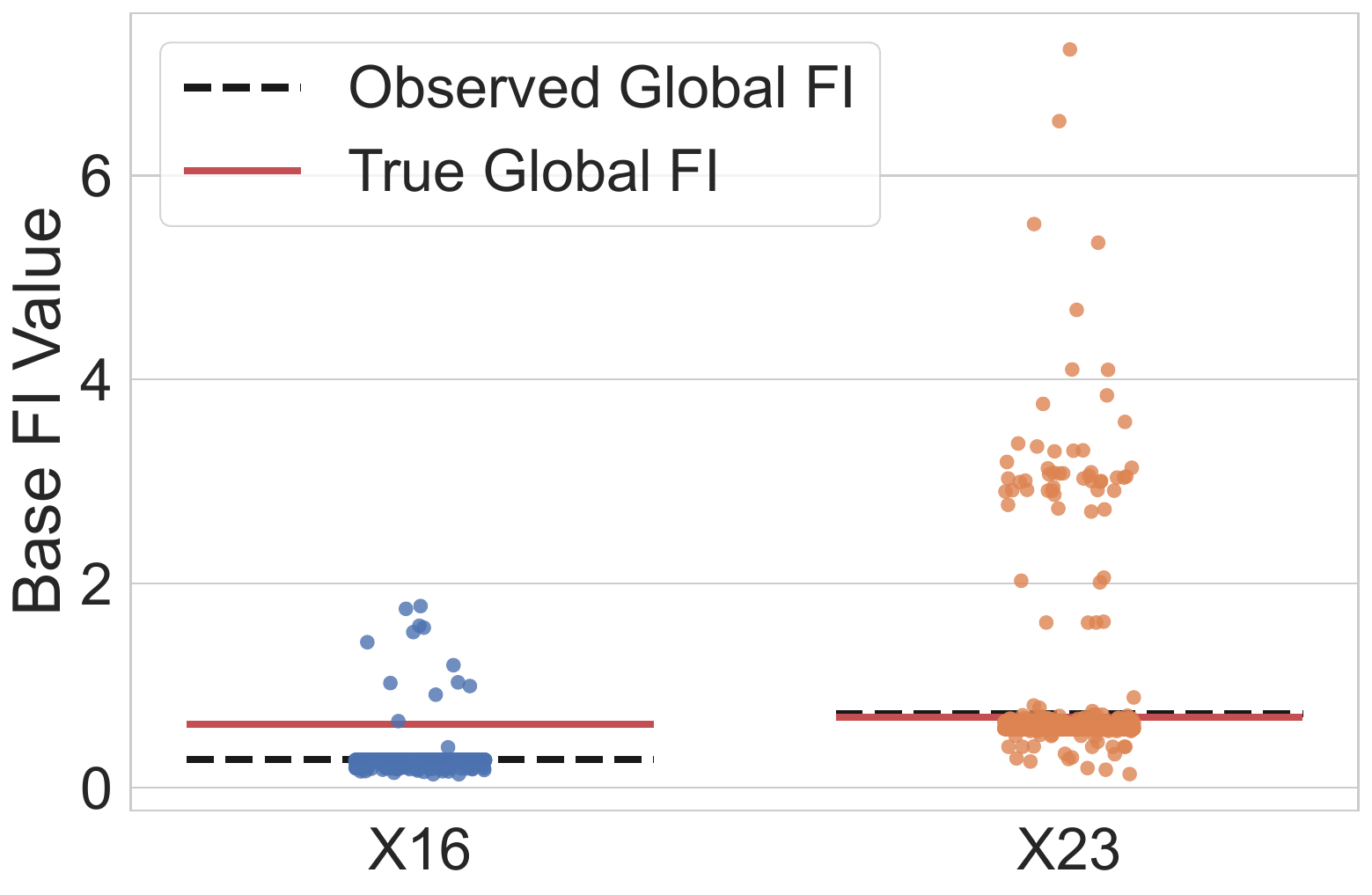}
    \caption[]%
     {Base FI values' distributions for features $X16$ and $X23$.}
    \label{fig:non_normal_pair}
\end{figure}

We further analyze the CIs of the features for $p=30$, an RF model, and a single explanation set of size $n=100,000$. In Figure \ref{fig:non_normal_pair}, the base FI values distribution of two features, for which we found coverage errors for multiple explanation sets. The true global FI values of the two features are almost identical, and the variance is relatively large. More importantly, the distributions of the base FI values of both features display extremely long tails, and the observed global FI values are influenced by the rare values at the tails. In such cases, we recommend replacing the paired t-test with a robust alternative \citep{wilcox2011introduction}.

\subsection{High-Dimensional Example}\label{app:high-dim}

\begin{figure}[ht]
    \centering\includegraphics[width=\columnwidth]{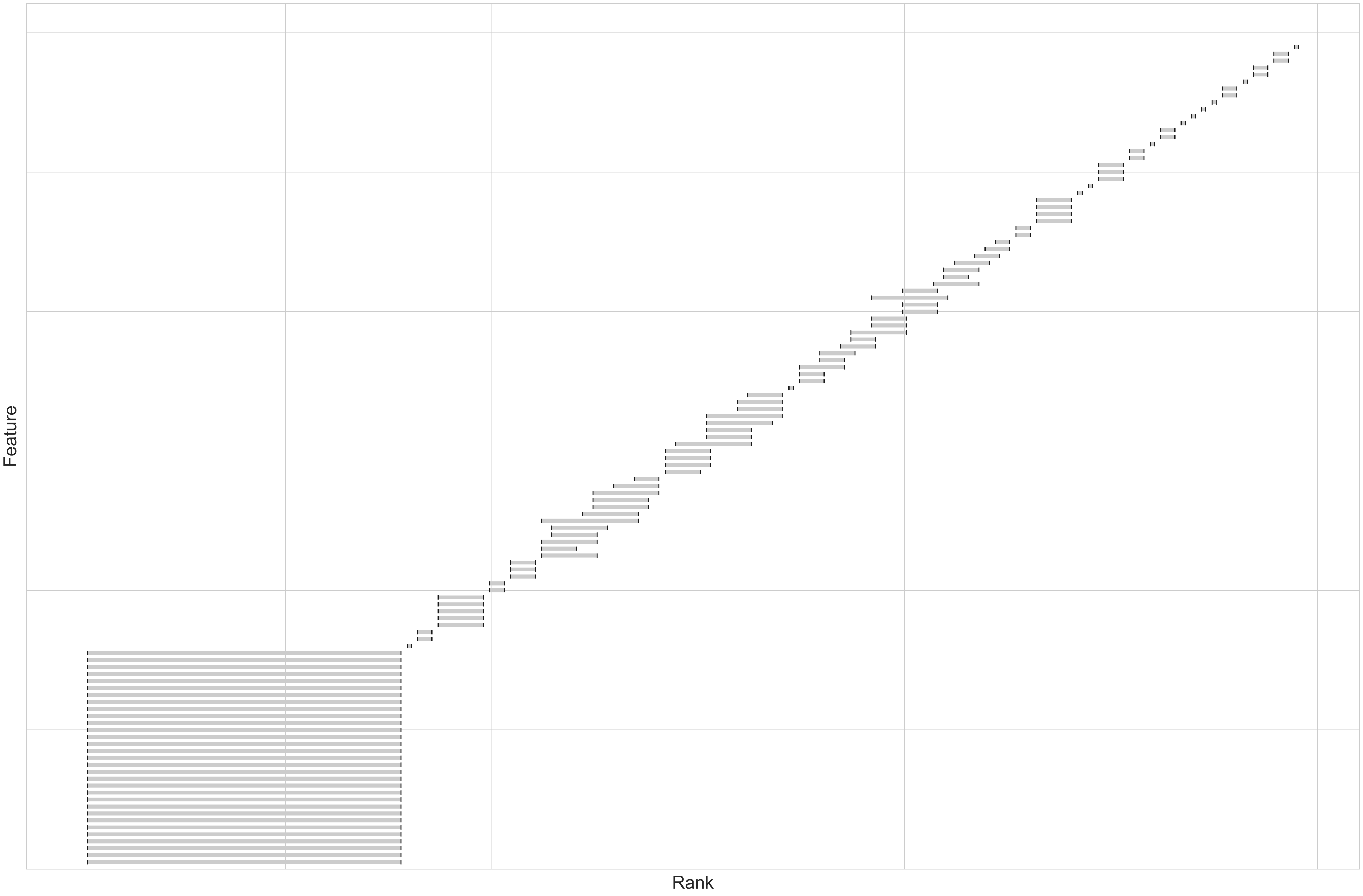}
    \caption[]%
     {CIs for the true ranks of the Nomao dataset features. The features are ordered by their observed global FI value. There are 31 irrelevant features and many intersections between CIs.}
    \label{fig:high_dim_ranking}
\end{figure}

Displaying the global FI values or the CIs for the ranks for many features may be difficult to interpret. Therefore, typically only the top-k features are presented. For example, in SHAP's global bar plot API\footnote{\href{https://shap.readthedocs.io/en/latest/generated/shap.plots.bar.html}{Global bar plot API}.} the default number of features to present is 10. In Figure \ref{fig:high_dim_ranking}, we present the complete ranking for the Nomao dataset\citep{misc_nomao_227} features . Our ranking method makes it easier to interpret which features are irrelevant and determine how to select a threshold $k$ for the most important features to display.

\newpage
\section{Implementation Details}\label{app:imp_details}

\subsection{Availability of Data}

The bike sharing dataset \citep{fanaee2014event} contains 10,886 records of bike rentals between 2011 and 2012 from the Capital Bikeshare program in Washington, D.C. The regression task is forecasting demand for bike rentals based on time and environmental measures such as month and weather. The data is publicly available at \href{https://www.kaggle.com/competitions/bike-sharing-demand/data}{Kaggle}.

The COMPAS dataset \citep{angwin2016machine} contains 6,172 records of criminal history from Broward County from 2013 and 2014. The classification task is assessing a criminal defendant’s likelihood to re-offend based on jail and prison time, and demographics. The data is publicly available at \href{https://github.com/propublica/compas-analysis}{Propublica's Github}.

Nomao is a search engine of places. The Nomao dataset \citep{misc_nomao_227} contains 34,465 records with comparison features about places, such as name and localization, from different sources. The classification task is detecting whether two sources of information refer to the same place. The data is publicly available at \href{https://www.openml.org/search?type=data&sort=runs&id=1486&status=active}{OpenML}.

\subsection{Reproducibility Instructions}

The code for our ranking method, experiments, and visualizations was written in the Python programming language (Python version 3.10.3) and can be found in this \href{https://github.com/BityaNeuhof/confident_feature_ranking}{Git repository}.  The ICRanks R package was imported to Python through the rpy2 package.

\subsection{Computing Infrastructure}

The experiments were performed on a server with 64G RAM and 16 CPUs.

\end{document}